\PassOptionsToPackage{table}{xcolor}
\documentclass[11pt]{article}

\usepackage[preprint]{acl}

\usepackage{times}
\usepackage{latexsym}

\usepackage[T1]{fontenc}

\usepackage[utf8]{inputenc}

\usepackage{microtype}

\usepackage{inconsolata}

\usepackage{times}
\usepackage{microtype}
\usepackage{url}
\usepackage{hyperref}

\usepackage{algorithm}
\usepackage{algorithmic}

\usepackage{amsmath}
\usepackage{amssymb}

\usepackage{amsmath}
\usepackage{amssymb}
\usepackage{mathtools}
\usepackage{amsthm}

\usepackage{graphicx}
\usepackage{subcaption}
\usepackage{caption}
\usepackage{tikz}
\usepackage[most]{tcolorbox}

\usepackage{array}
\usepackage{booktabs}
\usepackage{tabularx}
\usepackage{multirow}
\usepackage{xcolor}
\usepackage{colortbl}
\usepackage{cuted}
\usepackage{placeins}
\usepackage{dblfloatfix}

\definecolor{polarpurple}{RGB}{245,240,255}

\newcommand{\up}[1]{\ensuremath{_{\textcolor{red!75!black}{\scriptstyle #1}}}}
\newcommand{\down}[1]{\ensuremath{_{\textcolor{blue!75!black}{\scriptstyle #1}}}}
\newcommand{\bestres}[1]{\underline{\textbf{#1}}}

\newcolumntype{L}{>{\raggedright\arraybackslash}X}

\usepackage{cuted}
\usepackage{capt-of}
\usepackage{tabularx}
\usepackage{array}

\usepackage{enumitem}

\usepackage{pgfplots}
\pgfplotsset{compat=1.17}

\usepackage{enumitem}

\newlist{compactitem}{itemize}{1}
\setlist[compactitem,1]{
  label=\textbullet,
  leftmargin=1.6em,
  topsep=2pt,
  itemsep=1pt,
  parsep=0pt
}
\newcommand{\pcfield}[2]{\textbf{#1} & #2\\}
\newcommand{\pcsep}{\par\vspace{1pt}\noindent\rule{\linewidth}{0.3pt}\par\vspace{1pt}}
\newtcolorbox{skeletonbox}[1][]{
  enhanced,
  breakable,
  width=\linewidth,
  colback=black!2,
  colframe=black!25,
  boxrule=0.4pt,
  arc=2mm,
  left=1mm,
  right=1mm,
  top=0.75mm,
  bottom=0.75mm,
  #1
}

\newtcolorbox{promptcard}[2][]{
  enhanced,
  breakable,
  width=\linewidth,
  colback=white,
  colframe=black!50,
  boxrule=0.6pt,
  arc=2mm,
  left=1.2mm,
  right=1.2mm,
  top=1mm,
  bottom=1mm,
  fonttitle=\bfseries,
  fontupper=\scriptsize,
  title={#2},
  #1
}

\newtcolorbox{widepromptbox}[1]{
  enhanced,
  width=0.98\textwidth,
  colback=white,
  colframe=black!50,
  boxrule=0.6pt,
  arc=2mm,
  left=3mm,
  right=3mm,
  top=2mm,
  bottom=2mm,
  fonttitle=\bfseries,
  fontupper=\footnotesize,
  title={#1}
}

\newcounter{promptcard}
\renewcommand{\thepromptcard}{C\arabic{promptcard}}

\newenvironment{widepromptcard}[2]{%
  \refstepcounter{promptcard}\label{#1}%
  \begin{center}
  \centering
  \begin{widepromptbox}{Prompt Card~\thepromptcard: #2}
}{%
  \end{widepromptbox}
  \end{center}
}
\title{\textit{\textbf{PolarMem}}: A Training-Free Polarized Latent Graph Memory for Verifiable Vision-Language Models}

\author{
  \textbf{Zhisheng Chen\textsuperscript{1,2}$^*$},
  \textbf{Tingyu Wu\textsuperscript{1,2}$^*$},
  \textbf{Zijie Zhou\textsuperscript{3}$^*$},
  \textbf{Zhengwei Xie\textsuperscript{4}},
  \textbf{Jinhan Li\textsuperscript{4}},
  \textbf{Ziyan Weng\textsuperscript{5}},\\
  \textbf{Liang Lin\textsuperscript{2}},
  \textbf{Jingwei Song\textsuperscript{6}},
  \textbf{Zikai Xiao\textsuperscript{7}}$^\dagger$,
  \textbf{Yingwei Zhang\textsuperscript{1,2}}$^\dagger$\\
  \textsuperscript{1}ICT, CAS, 
  \textsuperscript{2}UCAS,
  \textsuperscript{3}CUPB,
  \textsuperscript{4}USTC,
  \textsuperscript{5}CityU-DG,
  \textsuperscript{6}HKU,
  \textsuperscript{7}ZJU
  \\
  \textsuperscript{$*$}Equal contribution \quad 
  \textsuperscript{$\dagger$}Corresponding authors
}

\begin{document}
\maketitle

\begin{figure*}[!t]
\centering
\includegraphics[
    width=0.96\textwidth,
    height=0.34\textheight,
    keepaspectratio
]{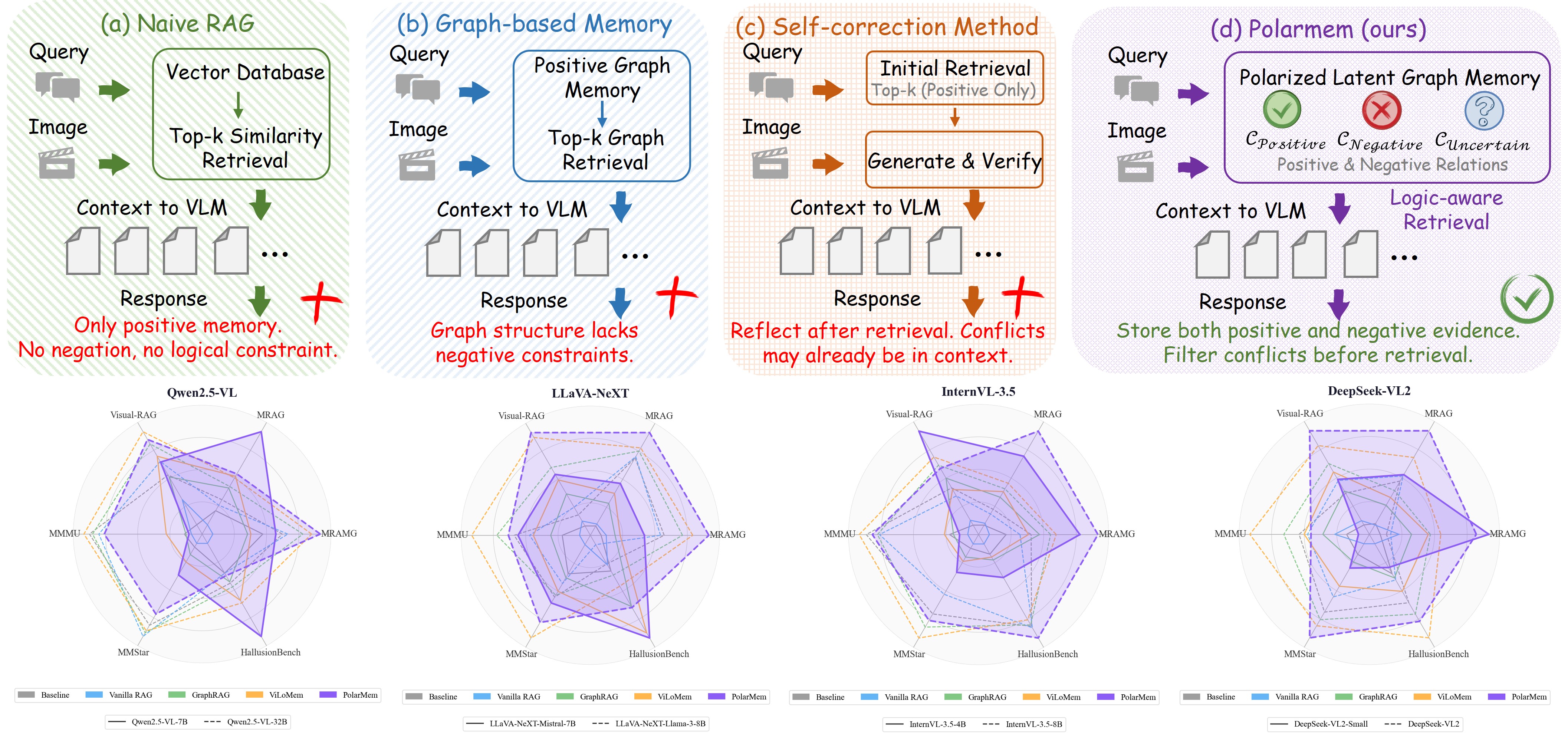}
\caption{
\textbf{Comparison between PolarMem and existing memory-based reasoning paradigms.}
The top row illustrates the key methodological differences among Naive RAG, graph-based memory, self-correction methods, and PolarMem. Naive RAG relies on top-$k$ similarity retrieval from a vector database and lacks explicit negation modeling. Graph-based memory improves structural organization but still mainly stores positive relations. Self-correction methods verify retrieved evidence after it has already entered the context, so conflicting memories may still influence generation. In contrast, PolarMem constructs a polarized latent graph memory with positive, negative, and uncertain states, and performs logic-aware retrieval to filter conflicting evidence before context construction. The bottom row summarizes performance across representative VLM backbones and multimodal benchmarks, showing that PolarMem provides more reliable retrieval-augmented reasoning ability.
}
\label{fig:overview}
\end{figure*}

\begin{abstract}
Memory is not merely a storage mechanism for intelligent systems, but a structure for organizing evidence and constraining belief.
This is especially important for multimodal reasoning, where retrieved evidence must be both query-relevant and visually consistent.
However, current memory systems for vision-language models (VLMs) remain largely positive-associative: they retrieve what is similar or previously observed, but lack an explicit way to remember what has been verified as absent or logically excluded.
To this end, we propose \textbf{PolarMem}, a training-free polarized latent graph memory framework for verifiable vision-language reasoning. 
PolarMem transforms frozen VLM perceptual signals into \textit{HAS}, \textit{NOT\_HAS}, and \textit{Uncertain} memory states through semantic consistency verification and adaptive distributional partitioning, and stores them in a polarized graph with distinct positive and negative memory relations. 
During inference, a lexicographical logic-aware retrieval protocol enforces logical consistency before semantic similarity, suppressing conflicting memories before they enter the model context. 
Across eight frozen VLM backbones and six multimodal benchmarks, PolarMem consistently improves retrieval-intensive tasks and reduces retrieval-level contradictions. 
These results highlight negative memory as a key mechanism for building more reliable multimodal memory systems.
Our code is available at \url{https://github.com/czs-ict/PolarMem}.
\end{abstract}

\section{Introduction}
Memory enables intelligent systems to move beyond immediate reaction toward sustained reasoning by organizing beliefs, preserving evidence, revising judgments, and ruling out errors~\citep{wu2024longmemeval}. A reliable memory system should therefore record not only what has been observed, but also what has been excluded, which hypotheses have been rejected, and which similar experiences should not guide the current decision~\citep{maharana2024evaluating}. This role becomes increasingly important as LLMs and VLMs are extended into multimodal agents, where external memory supports reasoning beyond parametric priors~\citep{wang2024mementos,luo2026survey,gutierrez2024hipporag}. However, when memory systems cannot distinguish semantic similarity from factual validity, retrieved evidence may amplify hallucination and conflicts rather than improve reliability~\citep{niu2024ragtruth,xu2024knowledge}.

Existing multimodal memory methods, including long-context modeling, vector databases, and graph-structured memory, mainly improve memory coverage and accessibility~\citep{zhao2024longrag,li2024graphreader,guo2024lightrag,hou2026flashmem}. Yet they often exhibit a structural positive bias: they effectively represent what exists, what is relevant, and what is similar, but rarely encode what does not exist, which hypotheses have been ruled out, or which similar evidence conflicts with current facts. Many multimodal reasoning errors therefore stem not from the lack of relevant memory, but from retrieving semantically related yet logically contradictory evidence~\citep{chen2024benchmarking}. To support verifiable multimodal reasoning, memory should move from positive evidence storage toward polarized belief organization, preserving both visually supported and visually excluded content~\citep{guan2024hallusionbench}. Only when negation becomes an independent memory state can the system actively suppress similar but conflicting evidence, transforming memory from a passive repository into an active constraint structure for reasoning~\citep{he2025evaluating}.

Motivated by this idea, we propose \textbf{PolarMem}, a training-free polarized graph memory framework for verifiable multimodal memory. PolarMem does not train a new vision-language model; instead, it reorganizes the perceptual signals of a frozen VLM at inference time and converts ambiguous visual confidence into memory states with logical constraint capabilities. Specifically, PolarMem first extracts candidate concepts from visual inputs and estimates their confidence through multi-prompt semantic consistency verification. It then applies adaptive distribution partitioning to classify concepts into three states: \textbf{HAS}, \textbf{NOT HAS}, and \textbf{Uncertain}, corresponding to evidence-supported existence, evidence-excluded negation, and unresolved uncertainty. These states are stored in a polarized graph consisting of visual nodes, textual nodes, and concept nodes. Unlike conventional graph memory, which mainly relies on positive semantic edges, PolarMem introduces both \textbf{HAS} and \textbf{NOT HAS} edges, making negative knowledge an explicit memory constraint rather than treating it as low similarity or missing information~\citep{leng2024mitigating}.

During retrieval, PolarMem reformulates memory reading from similarity maximization into logic-prioritized retrieval. Given a query, the system first parses target concepts and avoidance constraints, then checks whether each candidate memory violates the \textbf{HAS} / \textbf{NOT HAS} relations in the polarized graph. Semantic similarity is used for ranking only after logical consistency is satisfied~\citep{saad2024ares}. This lexicographic retrieval strategy prevents logical consistency and semantic relevance from being collapsed into a single continuous score, ensuring that negation constraints take priority. Thus, even highly similar memories are suppressed if they conflict with the current visual facts.

We evaluate PolarMem on multiple multimodal retrieval-augmented and visual reasoning benchmarks with frozen VLMs of different scales and architectures~\citep{yu2025visrag}. Experiments show that PolarMem consistently improves retrieval-intensive tasks, especially when models must distinguish supporting evidence from conflicting evidence in external memory. Comparisons with Vanilla RAG, GraphRAG-style baselines, and existing multimodal memory methods show that increasing memory capacity or adding graph structure alone is insufficient to resolve logical conflicts. Explicitly modeling \textbf{NOT HAS} constraints and enforcing logic-prioritized retrieval are key to improving multimodal memory verifiability. We also observe a trade-off between verifiability and reasoning flexibility under strict constraints, especially for stronger models or open-ended reasoning tasks, and therefore further analyze Hard, Soft, and Adaptive variants of PolarMem.

This paper makes the following contributions:

\begin{itemize}
    \item \textbf{We reveal the need for explicit negative memory in multimodal memory.}
    We show that existing systems are biased toward positive and similar evidence, while lacking explicit modeling of negation, exclusion, and conflicts.

    \item \textbf{We propose a training-free polarized graph memory framework.}
    PolarMem converts frozen VLM signals into \textbf{HAS}, \textbf{NOT HAS}, and \textbf{Uncertain} states, and stores positive evidence and negative constraints in a unified polarized graph.

    \item \textbf{We introduce logic-prioritized memory retrieval.}
    PolarMem enforces \textbf{HAS} / \textbf{NOT HAS} consistency before semantic ranking, preventing conflicting memories from entering the VLM context.

    \item \textbf{We evaluate both effectiveness and limitations.}
    Experiments show that explicit negative memory improves verifiability, while also revealing the effects of concept coverage, graph redundancy, and indexing cost.
\end{itemize}

\section{Related Work}

\textbf{Multimodal Agent Memory Systems.} The transition from stateless Large Language Models (LLMs) to persistent agents has necessitated the development of robust memory architectures, categorized recently by form into token-level, parametric, and latent memory systems \citep{hu2025memory}. Early approaches primarily relied on flattening multimodal interaction histories into linear token sequences \citep{zhong2024memorybank, packer2023memgpt}, which often suffer from information loss. To address this, recent works have adopted structured representations. \textbf{M3-Agent} \citep{long2025seeing}introduces an entity-centric memory graph that unifies visual and auditory modalities, explicitly linking face and voice identities to support long-horizon streaming interactions. Similarly, \textbf{KARMA}\citep{wang2025karma} employs a dual-memory architecture for embodied agents, utilizing a hierarchical 3D scene graph for long-term spatial grounding alongside a volatile short-term buffer for dynamic object states. In the navigation domain, \textbf{Mem2Ego}\citep{zhang2025mem2ego} projects global semantic maps into ego-centric views, enabling agents to leverage historical spatial data for immediate decision-making. These graph-based methods enhance structural organization but rely primarily on positive associations.

\textbf{Dual-Stream and Latent Cognitive Architectures.} The converging trend is the adoption of dual-stream architectures to decouple perception from reasoning. \textbf{ViLoMem}\citep{bo2025agentic} explicitly separates memory into a visual spoke for distraction patterns and a logic spoke for reasoning errors, preventing perceptual hallucinations from cascading into logical failures. In the latent space, \textbf{VisMem}\citep{yu2025vismem} overcomes the visual processing bottleneck by maintaining distinct short-term and long-term latent memories, synthesizing continuous memory tokens directly into the generation stream. Similarly, \textbf{MemoryVLA}\citep{shi2025memoryvla} fuses high-level cognitive tokens with low-level perceptual features to maintain temporal coherence in robotic manipulation. While these systems improve modularity, they rely on retrieving similar past instances.

\textbf{Inference-Time Interventions and Hallucination Mitigation.} Addressing hallucinations during inference without retraining has become a critical research domain. \textbf{Visual Inference-Time Intervention}\citep{sun2025v} identifies visual neglect by monitoring head-level activations and intervenes only when the model fails to prioritize input images. Similarly, \textbf{Robust Contrastive Decoding}\citep{chen2025decoupling} attempts to rectify confidence distributions by contrasting logits against plausible hallucination patterns. However, these methods operate primarily at the level of attention maps or logits.

\section{Methodology}
\label{sec:methodology}

\begin{figure*}[ht]
    \centering
    \includegraphics[width=\linewidth]{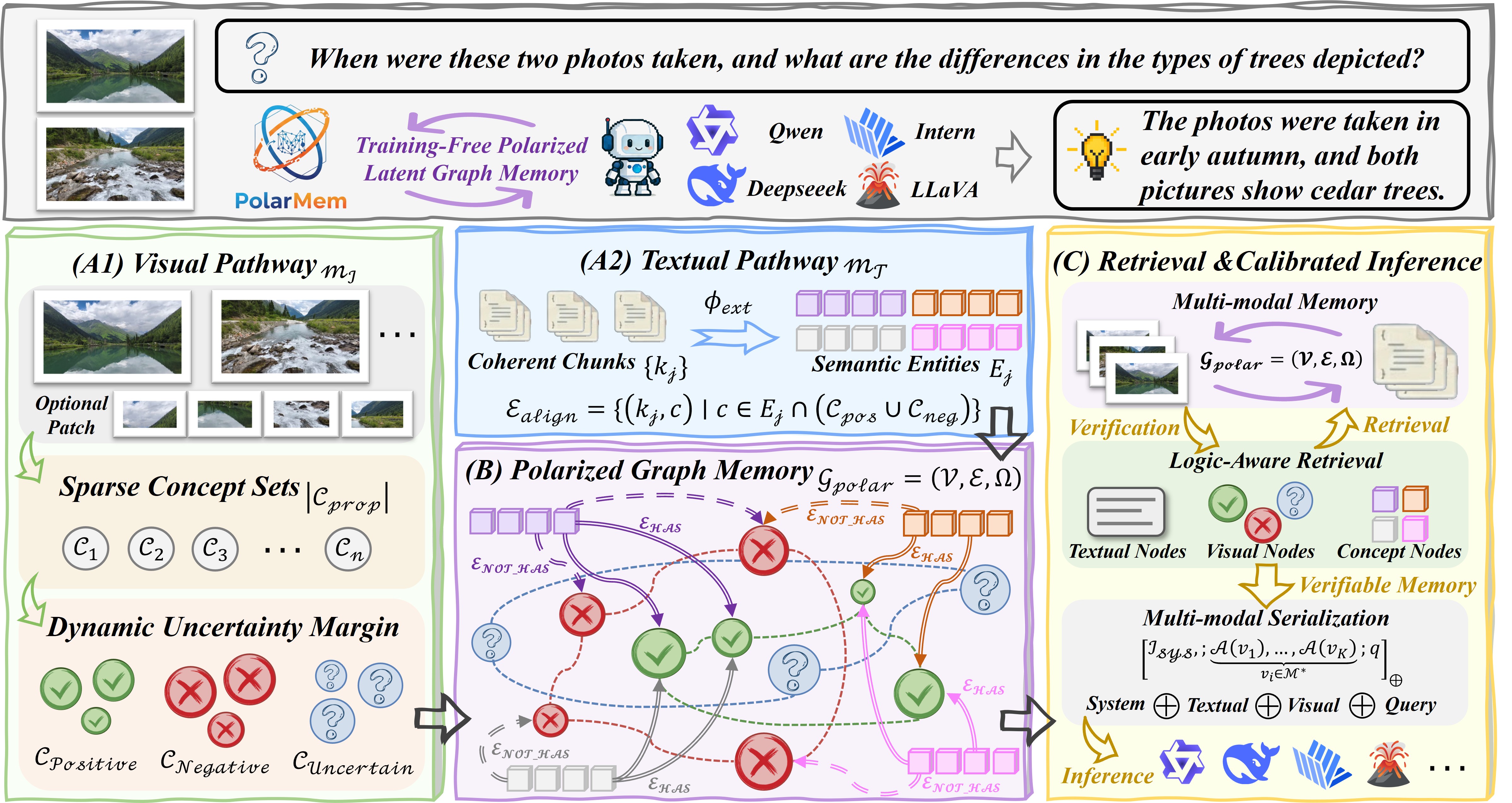}
    \caption{\textbf{Architectural overview of the PolarMem framework.}
    \textbf{(A) Dual-Pathway Logic Construction}: The \textit{Visual Pathway} ($m_I$) employs ensemble semantic consistency verification and adaptive distributional partitioning to categorize candidate concepts into \emph{Positive}, \emph{Negative}, and \emph{Uncertain} states. The \textit{Textual Pathway} ($m_T$) extracts semantic entities to establish deterministic alignment edges. 
    \textbf{(B) Polarized Latent Graph Memory ($\mathcal{G}_{polar}$)}: A heterogeneous topology that explicitly encodes negative knowledge via orthogonal $\mathcal{E}_{NOT\_HAS}$ edges, transforming fuzzy perceptual likelihoods into verifiable logical constraints. 
    \textbf{(C) Retrieval and Calibrated Inference}: A lexicographical logic-aware retrieval protocol prioritizes logical consistency over semantic similarity, ensuring constraint dominance. The resulting logically sanitized evidence is serialized into a multimodal context, enabling theVLM backbone to generate responses grounded in verifiable memory.}
    \label{fig:ablation_results}
\end{figure*}

We introduce PolarMem, a training-free polarized latent graph memory for verifiable multimodal retrieval. Rather than training a new vision-language model, PolarMem restructures the perceptual signals produced by a frozen backbone into explicit memory states. The key idea is to treat memory not only as a container of relevant evidence, but also as a constraint structure that records what is verified, what is ruled out, and what remains uncertain.

\subsection{Dual-Pathway Logic Construction}
\label{sec:logic_construction}

The first module converts raw multimodal inputs into concept-level logical states. It contains a visual pathway for probabilistic perception and a textual pathway for deterministic semantic alignment.

\paragraph{Visual pathway.}
For each visual episode $m_I$, PolarMem first uses the frozen VLM to propose an open-set candidate concept set $\mathcal{C}_{prop}$. These concepts are treated as hypotheses to be verified against the image. Since a single prompt can be sensitive to language priors and calibration noise, we use ensemble semantic consistency verification. Given a set of interrogation templates $\mathcal{T}_{ens}$, the verification score of a concept $c$ is computed as
\begin{equation}
\label{eq:calibrated_score}
s_c = \mathbb{E}_{\mathcal{T}\sim\mathcal{T}_{ens}}
\left[P_{\theta}(\text{``Yes''}\mid m_I,\mathcal{T}(c))\right].
\end{equation}
The resulting scores form an instance-specific confidence spectrum $\mathcal{S}_I=\{s_c\mid c\in\mathcal{C}_{prop}\}$.

To transform this continuous spectrum into memory states, we apply adaptive distributional partitioning. Rather than using a fixed threshold, PolarMem selects an image-specific boundary $\tau^*$ from the empirical score distribution. To stabilize the estimation under sparse or unimodal concept sets, we add anchor priors $s_{prior}\in\{0,1\}$ and maximize the inter-class variance:
\begin{equation}
\label{eq:variance_maximization}
\tau^* = \operatorname*{argmax}_{\tau\in[0,1]}
\left[\omega_{val}(\tau)\omega_{rej}(\tau)
(\mu_{val}(\tau)-\mu_{rej}(\tau))^2\right].
\end{equation}
We further introduce a dynamic uncertainty margin $\delta=\kappa\cdot\sigma_w$, where $\sigma_w$ is the weighted intra-class standard deviation. Candidate concepts are then partitioned into three states:
\begin{equation}
\label{eq:adaptive_partition}
\left\{
\begin{aligned}
    \mathcal{C}_{pos} &= \{c\mid s_c>\tau^*+\delta\},\\
    \mathcal{C}_{neg} &= \{c\mid s_c<\tau^*-\delta\},\\
    \mathcal{C}_{unc} &= \{c\mid |s_c-\tau^*|\leq\delta\}.
\end{aligned}
\right.
\end{equation}
Here, $\mathcal{C}_{pos}$ corresponds to verified positive evidence, $\mathcal{C}_{neg}$ corresponds to explicitly verified absence, and $\mathcal{C}_{unc}$ preserves ambiguous hypotheses without enforcing them as hard constraints.

\paragraph{Textual pathway.}
For textual memories $m_T$, we segment documents into coherent chunks $\{k_j\}$ and instantiate them as textual nodes. An entity extraction operator $\Phi_{ext}$ identifies semantic entities $E_j$ in each chunk. These entities establish alignment edges between textual memories and visual concept states:
\begin{equation}
\label{eq:text_alignment}
\mathcal{E}_{align}=\{(k_j,c)\mid c\in E_j\cap(\mathcal{C}_{pos}\cup\mathcal{C}_{neg})\}.
\end{equation}
This creates a shared concept space in which visual evidence and textual knowledge can be jointly addressed by the memory graph.

\subsection{Polarized Latent Graph Memory}
\label{sec:graph_topology}

The second module stores the constructed states in a heterogeneous graph
$\mathcal{G}_{polar}=(\mathcal{V},\mathcal{E},\Omega)$, where
$\mathcal{V}=\mathcal{V}_I\cup\mathcal{V}_T\cup\mathcal{V}_C$ contains visual nodes, textual nodes, and concept nodes. The key design is that negative knowledge is represented as a first-class memory relation rather than as low similarity or missing information. Specifically, visual-concept relations are polarized into two edge types:
\begin{equation}
\label{eq:edge_construction}
\begin{aligned}
\mathcal{E}_{\mathrm{HAS}} &= \{(v_I,v_c)\mid v_I\in\mathcal{V}_I,\; v_c\in\mathcal{C}_{pos}(v_I)\},\\
\mathcal{E}_{\mathrm{NOT\_HAS}} &= \{(v_I,v_c)\mid v_I\in\mathcal{V}_I,\; v_c\in\mathcal{C}_{neg}(v_I)\}.
\end{aligned}
\end{equation}
The $\mathcal{E}_{\mathrm{HAS}}$ edges store verified presence, while $\mathcal{E}_{\mathrm{NOT\_HAS}}$ edges store explicit negative constraints. This topology allows the memory to record both what the current evidence supports and what it rules out.

To support retrieval over both continuous semantics and discrete states, we define a hybrid embedding function $\Omega$. For each visual node $v_I$,
\begin{equation}
\label{eq:hybrid_embedding}
\Omega(v_I)=\left\langle \mathbf{z}_{\mathrm{vis}},\mathbf{Z}_{\mathrm{loc}},\mathbf{z}_{\mathrm{sem}}\right\rangle.
\end{equation}
Here, $\mathbf{z}_{\mathrm{vis}}$ captures holistic visual semantics, $\mathbf{Z}_{\mathrm{loc}}$ preserves local visual evidence, and $\mathbf{z}_{\mathrm{sem}}$ encodes the serialized polarized concept state:
\begin{equation}
\label{eq:concept_set_encoding}
\mathbf{z}_{\mathrm{sem}}=\mathrm{Enc}_{T}\left(
\mathcal{T}_{\mathrm{serialize}}(\mathcal{C}_{pos}(v_I),\mathcal{C}_{neg}(v_I))
\right).
\end{equation}
By encoding symbolic memory states back into latent space, $\Omega$ enables high-recall semantic matching while preserving explicit HAS and NOT\_HAS constraints for later logical filtering.

\subsection{Retrieval and Calibrated Inference}
\label{sec:retrieval_inference}

The third module reads from the polarized graph and constructs a verified context for the frozen VLM. Given a query $q$, PolarMem first parses it into target concepts $\mathcal{Q}^+$ and avoidance constraints $\mathcal{Q}^-$. It then evaluates each memory node by two criteria: a logical state $s_{\mathrm{log}}\in\{-1,0,1\}$ induced by the polarized graph, and a semantic score $s_{\mathrm{sem}}$ computed in the hybrid embedding space.

\begin{algorithm}[tb]
\caption{Lexicographical Logic-Aware Retrieval}
\label{alg:lexical_retrieval}
\small
\begin{algorithmic}[1]
\REQUIRE Query $q$, memory graph $\mathcal{G}=(\mathcal{V},\mathcal{E},\Omega)$, top-$K$
\ENSURE Retrieved context set $\mathcal{M}^*$
\STATE Parse $q$ into target concepts $\mathcal{Q}^+$ and avoidance constraints $\mathcal{Q}^-$
\STATE $\mathcal{L}\leftarrow\emptyset$
\FOR{each memory node $v\in\mathcal{V}$}
    \STATE $\mathbf{z}_v\leftarrow\Omega(v)$
    \STATE $s_{\mathrm{sem}}\leftarrow\mathrm{CosineSim}(\mathrm{Enc}(q),\mathbf{z}_v)$
    \STATE $\mathcal{C}^+(v)\leftarrow\{c\mid(v,c)\in\mathcal{E}_{\mathrm{HAS}}\}$
    \STATE $\mathcal{C}^-(v)\leftarrow\{c\mid(v,c)\in\mathcal{E}_{\mathrm{NOT\_HAS}}\}$
    \IF{$(\mathcal{Q}^+\cap\mathcal{C}^-(v)\neq\emptyset)\lor(\mathcal{Q}^-\cap\mathcal{C}^+(v)\neq\emptyset)$}
        \STATE $s_{\mathrm{log}}\leftarrow -1$ \COMMENT{Conflict}
    \ELSIF{$\mathcal{Q}^+\cap\mathcal{C}^+(v)\neq\emptyset$}
        \STATE $s_{\mathrm{log}}\leftarrow 1$ \COMMENT{Entailment}
    \ELSE
        \STATE $s_{\mathrm{log}}\leftarrow 0$ \COMMENT{Neutral}
    \ENDIF
    \STATE $\mathcal{L}.\mathrm{append}((s_{\mathrm{log}},s_{\mathrm{sem}},v))$
\ENDFOR
\STATE $\mathcal{L}^*\leftarrow\mathrm{SortDescending}(\mathcal{L},\mathrm{key}=\langle s_{\mathrm{log}},s_{\mathrm{sem}}\rangle)$
\STATE $\mathcal{M}^*\leftarrow\{v\mid(\cdot,\cdot,v)\in\mathcal{L}^*[:K]\}$
\RETURN $\mathcal{M}^*$
\end{algorithmic}
\end{algorithm}

As shown in Algorithm~\ref{alg:lexical_retrieval}, retrieval follows a lexicographical ranking rule. Memories that violate the query constraints are assigned $s_{\mathrm{log}}=-1$ and are ranked below logically compatible memories regardless of semantic similarity. Memories that provide verified positive evidence receive $s_{\mathrm{log}}=1$, while ambiguous or irrelevant memories receive $s_{\mathrm{log}}=0$. This avoids collapsing heterogeneous signals into a single weighted score such as $s_{\mathrm{sem}}+\lambda s_{\mathrm{log}}$ and instead enforces logical consistency as a retrieval priority.

The retrieved set $\mathcal{M}^*$ is serialized into a multimodal context. We define an assembly operator $\mathcal{A}$ that maps textual nodes to text and visual nodes to visual evidence tokens. The final context is
\begin{equation}
\label{eq:prompt_construction}
\mathbf{x}_{\mathrm{ctx}}=\left[\mathcal{I}_{\mathrm{sys}},\;\mathcal{A}(v_1),\ldots,\mathcal{A}(v_K),\;q\right]_{\oplus},
\quad v_i\in\mathcal{M}^*,
\end{equation}
where $[\cdot]_{\oplus}$ denotes sequential concatenation. The evidence order follows the lexicographical retrieval rank, so logically verified memories are placed before weaker or ambiguous evidence. The frozen VLM then generates the answer by conditioning on this verified context:
\begin{equation}
\label{eq:inference_objective}
\hat{y}=\operatorname*{argmax}_{y}\sum_{t=1}^{T}
\log P_{\theta_{\mathrm{frozen}}}(y_t\mid y_{<t},\mathbf{x}_{\mathrm{ctx}}).
\end{equation}
Thus, PolarMem does not update the backbone parameters; instead, it changes how memory is written, constrained, and read before generation.

\paragraph{Constraint strength.}
The lexicographical protocol above corresponds to the hard version of PolarMem. In experiments, we also analyze soft and adaptive variants to study the trade-off between verifiability and reasoning flexibility. The soft variant demotes conflicting memories with a penalty rather than categorically suppressing them, while the adaptive variant adjusts constraint strength according to retrieval confidence and query ambiguity. These variants retain the same polarized graph but differ in how strictly NOT\_HAS constraints are enforced during retrieval.

\paragraph{Boundary of training-free construction.}
Because PolarMem is training-free, its graph is bounded by the candidate concepts proposed and verified by the frozen VLM. If an answer-critical concept is never proposed in $\mathcal{C}_{prop}$, later graph construction cannot recover it. We therefore treat candidate proposal coverage as a measurable boundary of the current framework and analyze it empirically in the experiments.

\section{Experiments}

\begin{table*}[!t]
\centering
\caption{
Main results across eight frozen VLM backbones and six multimodal benchmarks.
Numbers in colored subscripts denote absolute changes relative to the baseline with the same backbone:
red indicates improvement and blue indicates decline.
Bold numbers indicate the best result within each backbone--benchmark block.
}
\scriptsize
\setlength{\tabcolsep}{3.4pt}
\renewcommand{\arraystretch}{1.12}
\resizebox{\textwidth}{!}{%
\begin{tabular}{@{}llcccccc@{}}
\toprule
\textbf{Backbone} & \textbf{Method}
& \textbf{MRAMG} & \textbf{MRAG} & \textbf{Visual-RAG}
& \textbf{MMMU} & \textbf{MMStar} & \textbf{HallusionBench} \\
\midrule

\multirow{5}{*}{\textbf{Qwen2.5-VL-7B}}
& Baseline     & 26.7 & 60.7 & 40.2 & 56.3 & 63.6 & 52.7 \\
& Vanilla RAG  & 22.1\down{-4.6} & 58.9\down{-1.8} & 44.1\up{+3.9} & 55.4\down{-0.9} & 63.4\down{-0.2} & 50.3\down{-2.4} \\
& GraphRAG     & 25.3\down{-1.4} & 63.6\up{+2.9} & 47.6\up{+7.4} & 56.1\down{-0.2} & 64.0\up{+0.4} & 53.4\up{+0.7} \\
& ViLoMem      & 25.6\down{-1.1} & 65.1\up{+4.4} & \bestres{50.8}\up{+10.6} & \bestres{58.6}\up{+2.3} & 64.4\up{+0.8} & 54.9\up{+2.2} \\
\rowcolor{polarpurple}
\cellcolor{white} & PolarMem     & \bestres{27.9}\up{+1.2} & \bestres{70.8}\up{+10.1} & 49.9\up{+9.7} & 55.7\down{-0.6} & \bestres{65.1}\up{+1.5} & \bestres{57.8}\up{+5.1} \\

\midrule
\multirow{5}{*}{\textbf{Qwen2.5-VL-32B}}
& Baseline     & 28.5 & 62.1 & 47.9 & 68.4 & 67.8 & 53.5 \\
& Vanilla RAG  & 29.0\up{+0.5} & 61.4\down{-0.7} & 50.0\up{+2.1} & 67.2\down{-1.2} & \bestres{68.4}\up{+0.6} & 52.7\down{-0.8} \\
& GraphRAG     & 30.4\up{+1.9} & 63.8\up{+1.7} & 52.7\up{+4.8} & 68.0\down{-0.4} & 68.2\up{+0.4} & 53.8\up{+0.3} \\
& ViLoMem      & 31.2\up{+2.7} & 64.9\up{+2.8} & \bestres{54.6}\up{+6.7} & \bestres{69.1}\up{+0.7} & 68.1\up{+0.3} & \bestres{55.1}\up{+1.6} \\
\rowcolor{polarpurple}
\cellcolor{white} & PolarMem     & \bestres{32.0}\up{+3.5} & \bestres{65.4}\up{+3.3} & 53.4\up{+5.5} & 66.5\down{-1.9} & 67.2\down{-0.6} & 52.9\down{-0.6} \\

\midrule
\multirow{5}{*}{\textbf{InternVL-3.5-4B}}
& Baseline     & 28.4 & 56.8 & 37.8 & 66.5 & 65.1 & 44.8 \\
& Vanilla RAG  & 27.2\down{-1.2} & 57.0\up{+0.2} & 38.1\up{+0.3} & 64.7\down{-1.8} & 64.9\down{-0.2} & 43.4\down{-1.4} \\
& GraphRAG     & 30.8\up{+2.4} & 59.8\up{+3.0} & 40.7\up{+2.9} & 66.1\down{-0.4} & 65.5\up{+0.4} & 45.6\up{+0.8} \\
& ViLoMem      & 30.0\up{+1.6} & 60.4\up{+3.6} & 40.0\up{+2.2} & \bestres{67.0}\up{+0.5} & 65.7\up{+0.6} & 45.2\up{+0.4} \\
\rowcolor{polarpurple}
\cellcolor{white} & PolarMem     & \bestres{33.7}\up{+5.3} & \bestres{64.2}\up{+7.4} & \bestres{43.6}\up{+5.8} & 65.5\down{-1.0} & \bestres{66.2}\up{+1.1} & \bestres{48.0}\up{+3.2} \\

\midrule
\multirow{5}{*}{\textbf{InternVL-3.5-8B}}
& Baseline     & 30.2 & 58.5 & 40.1 & 74.7 & 68.1 & 54.7 \\
& Vanilla RAG  & 29.4\down{-0.8} & 58.0\down{-0.5} & 39.6\down{-0.5} & 73.4\down{-1.3} & 67.2\down{-0.9} & 54.9\up{+0.2} \\
& GraphRAG     & 31.6\up{+1.4} & 60.7\up{+2.2} & 41.5\up{+1.4} & 73.6\down{-1.1} & 68.7\up{+0.6} & 54.5\down{-0.2} \\
& ViLoMem      & 32.0\up{+1.8} & 61.3\up{+2.8} & \bestres{42.0}\up{+1.9} & \bestres{75.4}\up{+0.7} & \bestres{69.2}\up{+1.1} & 53.8\down{-0.9} \\
\rowcolor{polarpurple}
\cellcolor{white} & PolarMem     & \bestres{35.0}\up{+4.8} & \bestres{66.9}\up{+8.4} & 41.3\up{+1.2} & 74.1\down{-0.6} & 68.4\up{+0.3} & \bestres{56.3}\up{+1.6} \\

\midrule
\multirow{5}{*}{\textbf{DeepSeek-VL2-Small}}
& Baseline     & 21.5 & 48.2 & 35.2 & 48.3 & 56.9 & 43.4 \\
& Vanilla RAG  & 22.4\up{+0.9} & 48.0\down{-0.2} & 35.6\up{+0.4} & 49.0\up{+0.7} & 56.2\down{-0.7} & 40.5\down{-2.9} \\
& GraphRAG     & 23.0\up{+1.5} & 51.1\up{+2.9} & 38.4\up{+3.2} & 49.7\up{+1.4} & 57.4\up{+0.5} & 43.8\up{+0.4} \\
& ViLoMem      & 23.8\up{+2.3} & 52.3\up{+4.1} & \bestres{40.2}\up{+5.0} & \bestres{50.8}\up{+2.5} & \bestres{59.0}\up{+2.1} & \bestres{45.1}\up{+1.7} \\
\rowcolor{polarpurple}
\cellcolor{white} & PolarMem     & \bestres{26.7}\up{+5.2} & \bestres{55.9}\up{+7.7} & 39.5\up{+4.3} & 47.6\down{-0.7} & 57.8\up{+0.9} & 42.8\down{-0.6} \\

\midrule
\multirow{5}{*}{\textbf{DeepSeek-VL2}}
& Baseline     & 21.8 & 54.9 & 38.2 & 51.1 & 60.7 & 46.2 \\
& Vanilla RAG  & 22.0\up{+0.2} & 55.7\up{+0.8} & 39.4\up{+1.2} & 50.6\down{-0.5} & 59.7\down{-1.0} & 44.0\down{-2.2} \\
& GraphRAG     & 23.9\up{+2.1} & 53.4\down{-1.5} & 41.0\up{+2.8} & 52.0\up{+0.9} & 61.2\up{+0.5} & 47.3\up{+1.1} \\
& ViLoMem      & 24.4\up{+2.6} & 58.6\up{+3.7} & 42.7\up{+4.5} & \bestres{54.0}\up{+2.9} & 61.6\up{+0.9} & \bestres{49.6}\up{+3.4} \\
\rowcolor{polarpurple}
\cellcolor{white} & PolarMem     & \bestres{26.1}\up{+4.3} & \bestres{62.8}\up{+7.9} & \bestres{44.1}\up{+5.9} & 50.4\down{-0.7} & \bestres{62.4}\up{+1.7} & 48.0\up{+1.8} \\

\midrule
\multirow{5}{*}{\textbf{LLaVA-NeXT-Mistral-7B}}
& Baseline     & 16.5 & 23.5 & 35.2 & 46.3 & 60.3 & 45.4 \\
& Vanilla RAG  & 17.2\up{+0.7} & 24.3\up{+0.8} & 35.9\up{+0.7} & 45.4\down{-0.9} & 58.8\down{-1.5} & 44.8\down{-0.6} \\
& GraphRAG     & 18.4\up{+1.9} & 32.8\up{+9.3} & 39.8\up{+4.6} & 46.9\up{+0.6} & 60.5\up{+0.2} & 48.9\up{+3.5} \\
& ViLoMem      & 19.1\up{+2.6} & 36.9\up{+13.4} & 41.8\up{+6.6} & 47.8\up{+1.5} & 61.2\up{+0.9} & 51.5\up{+6.1} \\
\rowcolor{polarpurple}
\cellcolor{white} & PolarMem     & \bestres{22.4}\up{+5.9} & \bestres{41.0}\up{+17.5} & \bestres{42.6}\up{+7.4} & \bestres{48.6}\up{+2.3} & \bestres{61.8}\up{+1.5} & \bestres{52.0}\up{+6.6} \\

\midrule
\multirow{5}{*}{\textbf{LLaVA-NeXT-Llama-3-8B}}
& Baseline     & 25.1 & 51.5 & 36.8 & 48.8 & 61.5 & 44.7 \\
& Vanilla RAG  & 24.6\down{-0.5} & 52.0\up{+0.5} & 37.9\up{+1.1} & 48.0\down{-0.8} & 60.8\down{-0.7} & 42.8\down{-1.9} \\
& GraphRAG     & 27.6\up{+2.5} & 54.3\up{+2.8} & 43.6\up{+6.8} & 49.7\up{+0.9} & 61.4\down{-0.1} & 45.8\up{+1.1} \\
& ViLoMem      & 29.0\up{+3.9} & 55.6\up{+4.1} & 47.9\up{+11.1} & \bestres{51.0}\up{+2.2} & \bestres{63.6}\up{+2.1} & 46.9\up{+2.2} \\
\rowcolor{polarpurple}
\cellcolor{white} & PolarMem     & \bestres{31.2}\up{+6.1} & \bestres{61.9}\up{+10.4} & \bestres{48.6}\up{+11.8} & 49.1\up{+0.3} & 62.8\up{+1.3} & \bestres{49.0}\up{+4.3} \\

\bottomrule
\end{tabular}%
}
\label{tab:main_results}
\end{table*}

\subsection{Experimental Setup}

\paragraph{Benchmarks.}
We evaluate PolarMem on six multimodal benchmarks covering retrieval-augmented reasoning, multimodal generation, general visual reasoning, and hallucination robustness:
\textbf{MRAMG-Bench}~\citep{yu2025mramg},
\textbf{MRAG-Bench}~\citep{hu2024mrag},
\textbf{Visual-RAG}~\citep{wu2025visual},
\textbf{MMMU}~\citep{yue2024mmmu},
\textbf{MMStar}~\citep{chen2024we},
and \textbf{HallusionBench}~\citep{guan2024hallusionbench}.
These benchmarks allow us to assess both retrieval-intensive settings and broader multimodal reasoning scenarios.

\paragraph{Backbones.}
We test eight frozen VLM backbones across different architectures and scales:
Qwen2.5-VL (7B / 32B), InternVL-3.5 (4B / 8B), DeepSeek-VL2 (Small / Base), and LLaVA-NeXT (Mistral-7B / Llama-3-8B).
All backbone parameters are kept fixed, and PolarMem operates strictly as a training-free inference-time memory module.

\paragraph{Baselines.}
We compare PolarMem with four settings:
\textbf{Baseline}, which performs direct inference without external memory;
\textbf{Vanilla RAG}, which retrieves from the same memory pool using dense vector similarity;
\textbf{GraphRAG}, which introduces graph-based evidence organization; and
\textbf{ViLoMem}~\citep{bo2025agentic}, a recent dual-stream multimodal memory method.
All memory-based methods use the same memory sources and the same Top-$K$ retrieval budget.

\subsection{Main Results}
\label{sec:main_results}

Table~\ref{tab:main_results} summarizes the main results across eight frozen VLM backbones and six benchmarks.
PolarMem achieves the most consistent gains on retrieval-intensive benchmarks, especially MRAG-Bench and MRAMG-Bench.
On MRAG-Bench, it obtains the best score for all eight backbones, with large improvements over direct inference, e.g., +10.1 on Qwen2.5-VL-7B, +8.4 on InternVL-3.5-8B, and +17.5 on LLaVA-NeXT-Mistral-7B.
These results show that explicit polarized memory is particularly effective when the task requires grounding answers in external multimodal evidence.

Compared with Vanilla RAG and GraphRAG, PolarMem remains stronger in most retrieval-heavy settings.
This indicates that the improvement is not merely due to retrieving more evidence or using a graph structure, but comes from explicitly encoding negative memory and enforcing logic-aware retrieval.
In contrast, standard similarity-based retrieval may introduce semantically related but factually conflicting evidence.

The results also reveal a clear boundary.
On general reasoning benchmarks such as MMMU and MMStar, PolarMem does not always outperform all baselines, especially on stronger backbones.
This suggests that strict logical constraints improve verifiability in evidence-grounded settings, but may reduce reasoning flexibility when the task relies more on the backbone's internal reasoning ability.
We analyze this trade-off in Section~\ref{sec:tradeoff}.

\subsection{Retrieval-Level Verifiability}
\label{sec:verifiability}

\begin{figure}[htbp]
  \centering
  \includegraphics[width=\columnwidth]{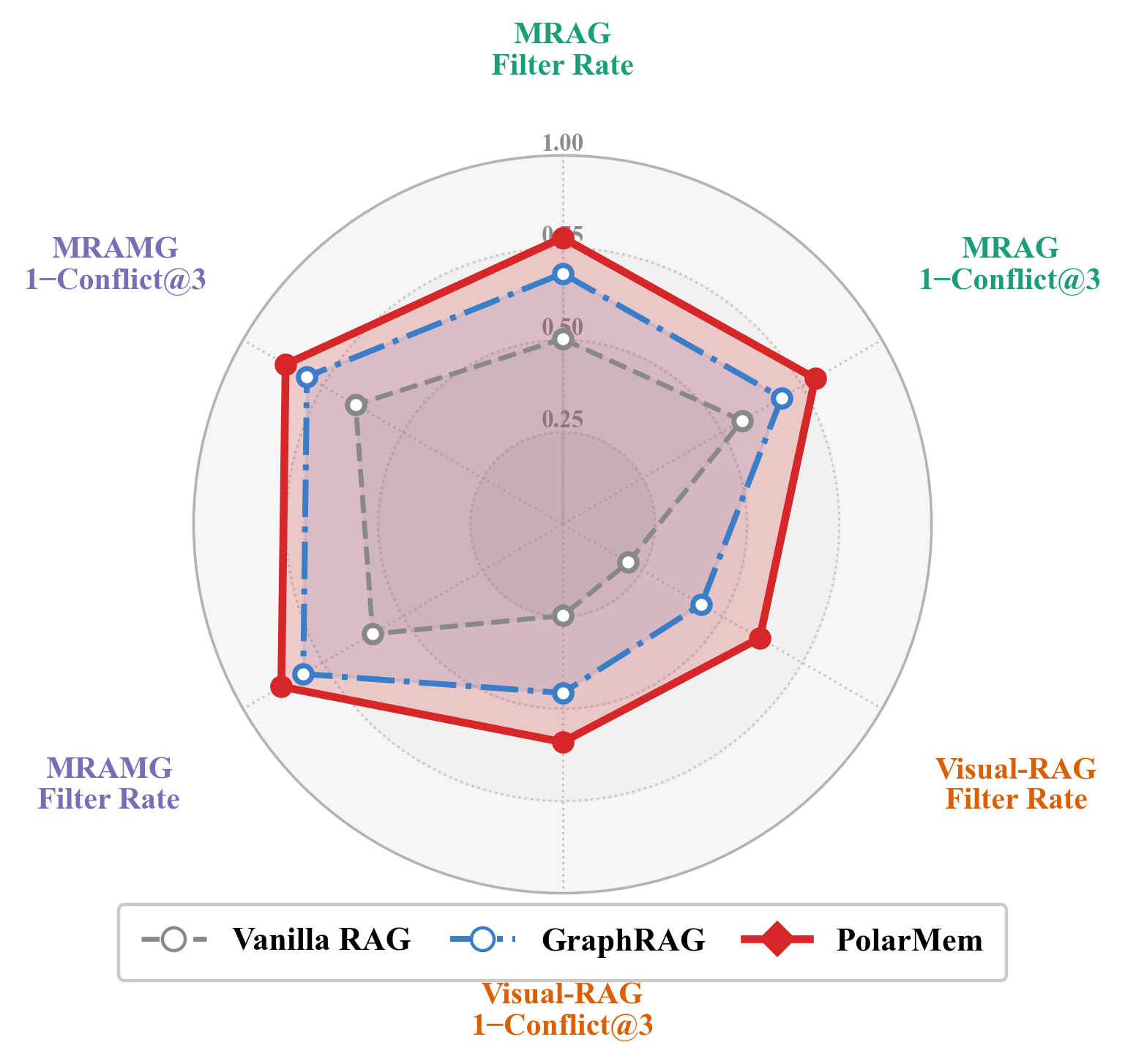}
  \caption{Retrieval-level contradiction filtering across retrieval-oriented benchmarks. Conflict@3 measures the fraction of contradictory evidence retained in the final top-3 context, while Filter Rate measures the fraction of contradictory candidates successfully excluded before context construction.}
  \label{fig:retrieval_verifiability}
\end{figure}

End-task performance does not directly show whether retrieved evidence is logically reliable.
We therefore evaluate retrieval-level verifiability on MRAG-Bench by measuring how well each method suppresses semantically similar but contradictory evidence.
We report \textbf{Conflict@3}, the proportion of contradictory evidence in the final top-3 retrieved results, and \textbf{Contradiction Filter Rate}, the fraction of contradictory candidates excluded from the final context.
Figure~\ref{fig:retrieval_verifiability} shows that PolarMem reduces Conflict@3 from 0.067 to 0.033 and improves the Filter Rate from 0.875 to 0.938.
This indicates that PolarMem improves verifiability at the retrieval stage by filtering logically conflicting memories before they enter the VLM context.

\begin{table*}[!t]
\centering
\small
\caption{
Sensitivity to candidate proposal coverage on Visual-RAG with Qwen2.5-VL-7B.
}
\setlength{\tabcolsep}{4.2pt}
\renewcommand{\arraystretch}{1.08}
\resizebox{\linewidth}{!}{
\begin{tabular}{lcccc}
\toprule
\textbf{Setting} 
& \textbf{Proposal Recall@K} 
& \textbf{Score (Covered)} 
& \textbf{Score (Missed)} 
& \textbf{Overall Score} \\
\midrule
Current proposer            & 83.2 & 55.0 & 38.1 & 49.9 {\scriptsize (+0.0)} \\
Multi-prompt proposer       & 89.6 & 55.3 & 38.9 & 51.4 {\scriptsize (+1.5)} \\
Global + local proposer     & 92.4 & \textbf{55.7} & \textbf{39.5} & 52.2 {\scriptsize (+2.3)} \\
Critical concept completion & \textbf{97.2} & 55.6 & --   & \textbf{53.1 {\scriptsize (+3.2)}} \\
\bottomrule
\end{tabular}
}
\label{tab:proposal_sensitivity}
\end{table*}

\subsection{Sensitivity to Candidate Proposal}
\label{sec:proposal_sensitivity}

PolarMem is training-free and therefore relies on the frozen VLM to propose candidate concepts before constructing HAS, NOT\_HAS, and Uncertain memory states.
To examine whether this proposal stage becomes a bottleneck, we evaluate several candidate proposal strategies on Visual-RAG with Qwen2.5-VL-7B.

Table~\ref{tab:proposal_sensitivity} shows that improving proposal coverage consistently improves the final score.
Multi-prompt and global-local proposal strategies increase Proposal Recall@K from 83.2 to 89.6 and 92.4, leading to gains of +1.5 and +2.3, respectively.
Critical concept completion further raises the score to 53.1, indicating that missed concepts are a real bottleneck.
However, the gain remains bounded, suggesting that verification, retrieval, and generation also affect the final performance.
These results clarify the boundary of PolarMem: it can organize and enforce constraints over proposed concepts, but cannot recover answer-critical concepts that are never proposed by the frozen VLM.

\subsection{Verifiability--Reasoning Flexibility Trade-off}
\label{sec:tradeoff}

\begin{table}[htbp]
\centering
\small
\caption{
Effect of different constraint strengths.
Hard retrieval is most effective on retrieval-intensive benchmarks, while Soft or Adaptive retrieval is more stable on general reasoning settings.
}
\setlength{\tabcolsep}{4.6pt}
\renewcommand{\arraystretch}{1.12}
\resizebox{\linewidth}{!}{%
\begin{tabular}{@{}llcccc@{}}
\toprule
\textbf{Benchmark} & \textbf{Backbone} 
& \textbf{Baseline} 
& \textbf{Soft} 
& \textbf{Adaptive} 
& \textbf{Hard} \\
\midrule

\multirow{2}{*}{\textbf{MRAMG}}
& Qwen2.5-VL-7B  
& 26.7 
& 26.9\up{+0.2} 
& 27.7\up{+1.0} 
& \bestres{27.9}\up{+1.2} \\
& Qwen2.5-VL-32B 
& 28.5 
& 30.1\up{+1.6} 
& 31.3\up{+2.8} 
& \bestres{32.0}\up{+3.5} \\

\midrule
\multirow{2}{*}{\textbf{MRAG}}
& Qwen2.5-VL-7B  
& 60.7 
& 63.4\up{+2.7} 
& 67.2\up{+6.5} 
& \bestres{70.8}\up{+10.1} \\
& Qwen2.5-VL-32B 
& 62.1 
& 62.5\up{+0.4} 
& 64.8\up{+2.7} 
& \bestres{65.4}\up{+3.3} \\

\midrule
\multirow{2}{*}{\textbf{MMMU}}
& Qwen2.5-VL-7B  
& 56.3 
& \bestres{56.7}\up{+0.4} 
& 56.6\up{+0.3} 
& 55.7\down{-0.6} \\
& Qwen2.5-VL-32B 
& \bestres{68.4} 
& 68.0\down{-0.4} 
& 68.3\down{-0.1} 
& 66.5\down{-1.9} \\

\midrule
\multirow{2}{*}{\textbf{HallusionBench}}
& Qwen2.5-VL-7B  
& 52.7 
& 56.6\up{+3.9} 
& 57.2\up{+4.5} 
& \bestres{57.8}\up{+5.1} \\
& Qwen2.5-VL-32B 
& 53.5 
& \bestres{53.9}\up{+0.4} 
& 53.8\up{+0.3} 
& 52.9\down{-0.6} \\

\bottomrule
\end{tabular}%
}
\label{tab:tradeoff}
\end{table}

Table~\ref{tab:tradeoff} shows that stricter logical constraints are most useful for retrieval-intensive tasks.
PolarMem-Hard achieves the best results on MRAMG and MRAG, where verified external evidence is central to answering.
However, on MMMU and stronger-backbone HallusionBench settings, Hard retrieval can be less stable, while Soft or Adaptive retrieval better preserves useful associative evidence.
This suggests a verifiability--reasoning flexibility trade-off: strict filtering improves evidence reliability, but excessive constraint strength may limit open-ended reasoning.

\paragraph{Summary.}
Overall, the experiments show that PolarMem is most effective when multimodal reasoning depends on reliable external evidence.
Its gains come from retrieval-level contradiction filtering rather than graph indexing alone.
Meanwhile, the trade-off and proposal-sensitivity analyses show that polarized memory is not a universal replacement for backbone reasoning:
strict logical constraints should be applied carefully, and the quality of candidate concept proposal remains an important boundary of the training-free setting.
We further report computational cost, graph redundancy, and single-episode construction details in Appendix~\ref{app:additional_analysis}.

\section{Conclusion}
We introduced PolarMem, a training-free polarized latent graph memory framework for verifiable vision-language reasoning. 
PolarMem moves multimodal memory beyond positive evidence storage by explicitly representing negative memory through NOT\_HAS constraints, together with HAS and Uncertain states derived from frozen VLM signals. 
By organizing these states in a polarized graph and enforcing logic-aware retrieval before semantic ranking, PolarMem suppresses semantically similar but logically conflicting evidence before it enters the VLM context. 
Experiments across eight frozen VLM backbones and six benchmarks show consistent gains on retrieval-intensive tasks and reduced retrieval-level contradictions. 
These findings suggest that explicit negative memory is a promising direction for building more reliable multimodal memory systems.

\clearpage
\section*{Limitations}
PolarMem focuses on training-free memory construction and logic-aware retrieval for verifiable vision-language reasoning. Since it does not update the underlying VLM, its memory quality depends on the perceptual signals and candidate concepts provided by the frozen backbone. Our current evaluation emphasizes retrieval-augmented and benchmark-based multimodal reasoning, while extending polarized memory to longer interactive settings with evolving memory states remains future work. PolarMem also introduces additional memory construction overhead compared with standard dense retrieval, though this cost can be amortized when the memory pool is reused. Future work may improve the framework through more efficient concept pruning, adaptive verification, and task-aware constraint calibration.


\section*{Ethics Statement}
This paper presents work whose goal is to advance the field of Machine Learning, specifically focusing on the reliability and safety of multimodal agents. By introducing a mechanism to explicitly suppress hallucinations and enforce logical consistency in long-horizon reasoning, our research contributes to the development of more trustworthy and verifiable AI systems. This is particularly critical as agents are increasingly deployed in decision-making roles where factual accuracy is paramount. We believe this work helps mitigate risks associated with probabilistic errors in large vision-language models. During manuscript preparation, large language models were used for language polishing, grammar correction, and improving the clarity of writing. 

\bibliography{custom}

\clearpage
\appendix
\sloppy

\begin{strip}
\vspace{-1.2em}
\centering
\captionof{table}{Overview of Multimodal Benchmarks used in evaluation. The datasets are categorized into General Understanding, Safety \& Robustness, and Multimodal RAG.}
\label{tab:benchmark_overview_app}

\small
\setlength{\tabcolsep}{4.0pt}
\renewcommand{\arraystretch}{1.22}

\begin{tabularx}{\textwidth}{@{} p{2.6cm} p{2.1cm} L p{2.1cm} @{}}
\toprule
\textbf{Benchmark} & \textbf{Core Type} & \textbf{Key Features \& Contributions} & \textbf{Data Scale} \\
\midrule

\textbf{MMMU} 
& General \newline Understanding 
& Covers 6 core disciplines (Art, Science, etc.) and 30 sub-fields. Evaluates expert-level AGI capabilities. 
& 11.5k \newline Questions \\

\textbf{MMStar} 
& Data Quality \newline (Dependency) 
& Addresses "visual redundancy" by selecting samples that strictly depend on visual information (Visual-Dependency). 
& Curated \newline Core Set \\

\textbf{HallusionBench} 
& Safety \newline \& Hallucination 
& Focuses on hallucination and visual illusion. Uses Control Groups to test logical consistency in reasoning. 
& 346 Images \newline 1129 Questions \\

\textbf{MRAG-Bench} 
& Retrieval \newline (Vision-Centric) 
& Identifies scenarios where visual retrieval outperforms text, e.g., multiple perspectives and occlusion. 
& 16k Images \newline 9 Scenarios \\

\textbf{MRAMG-Bench} 
& Multimodal \newline Generation 
& Focuses on the output side. Requires generating long-form answers containing both text and images. 
& 14k Images \newline 4.8k QA Pairs \\

\textbf{Visual-RAG} 
& Text-to-Image \newline Retrieval 
& Evaluates retrieving visual evidence, i.e., clue images, via text-to-image retrieval for knowledge-intensive QA. 
& Knowledge \newline Intensive \\

\bottomrule
\end{tabularx}

\end{strip}

\section{Detailed Overview of Multimodal Benchmarks}
\label{app:benchmark_details}

In this section, we provide a detailed overview of the six key multimodal benchmarks employed in our evaluation. These datasets cover a wide range of dimensions, from basic perception and expert-level reasoning to safety testing and Retrieval-Augmented Generation (RAG). Table~\ref{tab:benchmark_overview_app} summarizes their core features, and detailed descriptions follow below.

\subsection{General Understanding and Data Validity}

\textbf{MMMU (Massive Multi-discipline Multimodal Understanding).} 
MMMU sets the standard for measuring multimodal reasoning and understanding. Spanning 6 core disciplines (Art \& Design, Business, Science, Medicine, Humanities \& Social Sciences, and Engineering) and 30 sub-fields, it evaluates models on tasks demanding deep domain knowledge comparable to university-level exams.

\textbf{MMStar (Visual-Dependency Evaluation).}
MMStar was developed to rigorously audit the true multimodal capabilities of VLM, addressing the prevalence of visual redundancy and data leakage in prior benchmarks. Empirical analysis revealed that a significant portion of samples in existing datasets could be solved using textual commonsense or parametric knowledge alone, without processing the visual content. MMStar filters these out to create a curated touchstone dataset of samples that possess strict visual dependency, ensuring that successful answers genuinely reflect the model's ability to perceive and reason over visual content rather than relying on language priors.

\subsection{Safety and Hallucination Detection}

\textbf{HallusionBench.}
This benchmark diagnoses \emph{hallucinations} and \emph{visual illusions}. It employs a "Control Group" mechanism, analyzing consistency across logically related question pairs. It specifically tests robustness against visual illusions that often confuse even human perception.

\subsection{Multimodal Retrieval-Augmented Generation}

\textbf{MRAG-Bench (Vision-Centric RAG).}
MRAG-Bench focuses on the \emph{input utility} of visual retrieval. It defines 9 scenarios (e.g., temporal changes, different viewpoints) to verify when retrieving image evidence is more effective than text descriptions.

\textbf{MRAMG-Bench (Multimodal Generation).}
This benchmark targets the \emph{output modality}. Unlike traditional RAG which retrieves images to generate text, MRAMG requires the model to generate rich media answers (interleaved text and images), testing the ability to synthesize retrieved information.

\textbf{Visual-RAG (Text-to-Image Retrieval).}
Visual-RAG evaluates the \emph{acquisition} of visual evidence. It focuses on knowledge-intensive queries where the answer depends on finding specific "Clue Images" via text-to-image retrieval, testing the link between textual queries and visual verification.

\clearpage

\begin{strip}
\centering

\captionof{table}{
Computational cost comparison among Vanilla RAG, Pure GraphRAG, and PolarMem.
PolarMem introduces higher offline indexing cost due to concept verification and polarized graph construction, while the resulting memory can be reused across multiple queries.
}
\label{tab:cost_analysis}

\small
\setlength{\tabcolsep}{5pt}
\renewcommand{\arraystretch}{1.10}

\begin{tabular}{@{}llccc@{}}
\toprule
\textbf{Phase} & \textbf{Cost Metric} 
& \textbf{Vanilla RAG} 
& \textbf{Pure GraphRAG} 
& \textbf{PolarMem} \\
\midrule

\multirow{5}{*}{\textbf{Offline indexing}}
& Wall-clock time / episode (s) & 0.31 & 2.04 & 6.92 \\
& VLM/API calls / episode       & 1.00 & 2.00 & 23.97 \\
& Prompt tokens / episode       & 0    & 1946 & 23201 \\
& Output tokens / episode       & 0    & 43.4 & 38.8 \\
& Peak memory / episode         & 3142MB & 1658MB & 6518MB \\
\midrule

\multirow{3}{*}{\textbf{Online retrieval \& inference}}
& Retrieval latency / query (ms)             & 6.7 & 12.0 & 33.2 \\
& Logic-aware reranking latency / query (ms) & 143.9 & 45.4 & 118.1 \\
& End-to-end latency / query (ms)            & 3152.4 & 3179.9 & 6539.6 \\
\bottomrule
\end{tabular}

\captionof{table}{
Graph quality and retrieval efficiency across different backbones.
Redundancy ratio measures within-image duplicate or near-duplicate proposal rate and is not computed as $1-\text{Unique}/\text{Raw}$.
Inactive NOT\_HAS Ratio denotes the fraction of NOT\_HAS constraints that are not activated in Top-$K$ retrieval.
}
\label{tab:graph_quality}

\tiny
\setlength{\tabcolsep}{5pt}
\renewcommand{\arraystretch}{1}

\resizebox{\textwidth}{!}{%
\begin{tabular}{@{}lccc@{}}
\toprule
\textbf{Metric} 
& \textbf{InternVL-3.5-8B}
& \textbf{DeepSeek-VL2-Small}
& \textbf{LLaVA-NeXT-Mistral-7B} \\
\midrule

\multicolumn{4}{@{}l}{\textit{Concept proposal statistics}} \\
\midrule
Raw Concepts / Image 
& 29.46 & 64.34 & 70.26 \\
Unique Concepts / Image 
& 22.98 & 25.46 & 24.70 \\
Redundancy Ratio $\downarrow$ 
& \bestres{0.0693} & 0.2913 & 0.3723 \\
Overlap Ratio $\downarrow$ 
& \bestres{0.0302} & 0.0353 & 0.0817 \\
\midrule

\multicolumn{4}{@{}l}{\textit{Retrieval behavior}} \\
\midrule
Inactive NOT\_HAS Ratio $\downarrow$ 
& -- & 0.9813 & \bestres{0.9516} \\
Effective Concepts Top-$K$ $\uparrow$ 
& 2.535 & 2.825 & \bestres{4.855} \\
Evidence Yield / Concept $\uparrow$ 
& \bestres{1.5854} & 0.9102 & 0.5942 \\
\bottomrule
\end{tabular}%
}
\end{strip}

\section{Additional Experimental Analysis}
\label{app:additional_analysis}

This appendix provides additional analyses that complement the main experiments, including computational cost, graph quality, retrieval efficiency, and a fine-grained breakdown of single-episode memory construction.

\subsection{Computational Cost and Scalability}
\label{app:cost}

PolarMem introduces additional computation because it constructs polarized memory states before retrieval. 
To make this cost transparent, we compare Vanilla RAG, Pure GraphRAG, and PolarMem under the same memory source and retrieval budget. 
We separate the analysis into offline indexing and online retrieval/inference.

As shown in Table~\ref{tab:cost_analysis}, PolarMem is more expensive than Vanilla RAG, especially during offline indexing.
The major overhead comes from candidate verification and memory-state construction, rather than graph serialization itself.
However, this cost is incurred once for a reusable memory pool and can be amortized when the same memory is queried repeatedly.
Online retrieval is also slower than Vanilla RAG, but remains practical for benchmark-scale inference.
These results clarify that PolarMem trades additional memory construction cost for stronger retrieval-time verifiability.

\begin{table*}[!t]
\centering
\caption{
Single-episode cost breakdown for PolarMem memory construction.
Backend calls include verification-score access or other non-generative scoring operations; adaptive partitioning itself does not require natural-language generation.
}
\small
\setlength{\tabcolsep}{4pt}
\renewcommand{\arraystretch}{1.08}
\resizebox{\linewidth}{!}{
\begin{tabular}{lccccc}
\toprule
\textbf{Stage} 
& \textbf{Time (s)}
& \textbf{Backend Calls}
& \textbf{Prompt Tokens}
& \textbf{Output Tokens}
& \textbf{Peak Memory} \\
\midrule
Candidate concept generation       & 1.31 & 1.00  & 1247.8  & 27.6 & 1531MB \\
Ensemble verification              & 3.06 & 1.25  & 12634.0 & 0.0  & 4796MB \\
Adaptive partitioning              & 1.06 & 16.20 & 0.0     & 0.0  & 1521MB \\
Graph construction \& serialization & 0.09 & 0.00  & 0.0     & 0.0  & 1299MB \\
\midrule
\textbf{Total}                     & \textbf{5.53} & \textbf{18.45} & \textbf{13881.8} & \textbf{27.6} & \textbf{4796MB} \\
\bottomrule
\end{tabular}
}
\label{tab:single_episode_cost}
\end{table*}

\subsection{Graph Quality and Retrieval Efficiency}
\label{app:graph_quality}

We further analyze the quality and efficiency of the constructed concept graphs.
High concept coverage alone does not guarantee an efficient memory structure, since different backbones may generate redundant concepts or constraints that are rarely activated during retrieval.
We therefore report graph-side statistics and retrieval-side usage statistics across representative backbones.

Table~\ref{tab:graph_quality} shows that graph compactness is backbone-dependent.
InternVL-3.5-8B produces a relatively compact concept graph with low redundancy and high evidence yield.
In contrast, DeepSeek-VL2-Small and LLaVA-NeXT-Mistral-7B generate larger concept sets, but a smaller fraction of concepts is effectively used during Top-$K$ retrieval.
This suggests that explicit negative memory is useful, but its efficiency depends on the proposal behavior of the underlying backbone.
Future work may further improve PolarMem through concept normalization, pruning, and adaptive constraint activation.

\subsection{Single-Episode Cost Breakdown}
\label{app:single_episode_cost}

To identify the main source of PolarMem's overhead, we provide a fine-grained breakdown of memory construction for a single visual episode.
We decompose the process into candidate concept generation, ensemble verification, adaptive partitioning, and graph construction.

Table~\ref{tab:single_episode_cost} indicates that ensemble verification dominates the construction cost in terms of prompt tokens and peak memory.
Graph construction and serialization are lightweight by comparison.
This suggests that future efficiency improvements should primarily target verification reduction, such as selective concept verification, batched scoring, or pruning low-utility candidates before graph construction.

\clearpage
\section{Additional Experimental Details}
\label{app:exp_details}

\subsection{MRAMG Prompts and Evidence Templates}
\label{app:prompts}

To ensure fair comparison and reproducibility, we standardize evidence serialization and prompt interfaces across all memory-based methods. Each retrieved item is wrapped as \texttt{[Fact Check: \{Status\}] \{Content\}} and truncated to a fixed per-item budget before concatenation into the model context. We report the verbatim prompts used throughout the pipeline (offline indexing/concept extraction and online retrieval, coordinate-grounded answering, and candidate reranking), together with the corresponding system/developer instruction blocks and evidence formatting rules.

The full MRAMG templates are collected in the Prompt Card Catalog (Section~\ref{app:prompt_card_catalog}), with Prompt Cards~\ref{pc:mramg-figure-concept}--\ref{pc:mramg-reranking} corresponding to this pipeline.

\subsection{Qualitative Online Inference Examples}
\label{app:qual_examples}

We include two representative online inference traces to illustrate how retrieval, coordinate-grounded inspection, and evidence-constrained generation interact in practice. Both examples report the logged top-K retrieved document/image identifiers, the selected local patches, and the final answer together with overlap-based metrics, enabling transparent inspection of where performance gains and residual errors originate.

The two full traces are collected as Prompt Cards~\ref{pc:mramg-example-laser} and~\ref{pc:mramg-example-civrealm} in the Prompt Card Catalog.

\subsection{MRAG Prompts and Evidence Templates}
\label{app:mrag_prompts}

We use a standardized evidence template for all memory-based methods:
\texttt{[Fact Check: \{Status\}] \{Content\}}.
This appendix provides (i) the full prompts used at each pipeline stage, (ii) the system/developer instruction blocks when applicable, and (iii) evidence formatting and truncation rules.
All prompt cards follow a unified schema with explicit \texttt{Role}, \texttt{Inputs}, \texttt{Output}, \texttt{Gate}, \textbf{Hard Constraints}, and a verbatim \textbf{Skeleton}.

The full MRAG prompt templates are collected as Prompt Cards~\ref{pc:mrag-full-image-caption}--\ref{pc:mrag-rag-mcq} in the Prompt Card Catalog.

\subsection{Qualitative Inference Visualizations}
\label{app:qual_visualizations}
To make the end-to-end inference process transparent, we provide qualitative visualizations for representative test queries. Each example is rendered as a pair of prompt cards: (i) a \textbf{NoRAG (baseline)} card where the model answers using only the main query image, and (ii) a \textbf{RAG (full retrieval)} card where the same question is answered with the main image plus the top retrieved images as auxiliary visual evidence (and relationship context when enabled).
For each card, we display exactly what the model receives (images and the verbatim text prompt) as well as the observed output, enabling direct inspection of how retrieval augmentation changes the available evidence and the final decision.

The NoRAG and RAG visualizations are collected as Prompt Cards~\ref{pc:norag-q6} and~\ref{pc:rag-q6} in the Prompt Card Catalog.

\subsection{Prompt Card Catalog}
\label{app:prompt_card_catalog}

This catalog collects the full prompt cards referenced by Sections~\ref{app:prompts}--\ref{app:qual_visualizations}. Group headings are provided for navigation; individual card labels provide the cross-references used in the preceding subsections.

\clearpage
\onecolumn
\subsubsection{MRAMG Prompt Templates}

\begin{widepromptcard}{pc:mramg-figure-concept}{1. Offline: Figure Concept Extraction}
\begin{tabularx}{\linewidth}{@{}p{0.13\linewidth}X@{}}
\pcfield{Role}{Extract visual-verifiable concepts from a scientific figure for offline indexing.}
\pcfield{Inputs}{One figure image (global view), resized to max dimension 1024.}
\pcfield{Output}{Comma-separated list of short phrases (\(\le\)40).}
\pcfield{Gate}{N/A.}
\end{tabularx}

\pcsep
\textbf{Hard Constraints.}
\begin{itemize}[leftmargin=1.6em, topsep=1pt, itemsep=1pt, parsep=0pt]
  \item \textbf{Visual-verifiable only}: concepts must be observable from the figure.
  \item \textbf{Format}: output strictly as comma-separated short phrases; no numbering/bullets; no extra text.
  \item \textbf{Length}: max 40 items; avoid abstract/non-visual words.
\end{itemize}

\pcsep
\textbf{Skeleton.}
\begin{skeletonbox}
\ttfamily\footnotesize\raggedright\sloppy
[image]\par
This is a scientific figure (paper-style). List ONLY visual-verifiable concepts that appear in the figure, such as plot/diagram elements (axis, legend, line plot, bar chart, scatter, heatmap, table, pipeline, module, arrow, block, architecture diagram, flowchart, equation text, title text), and concrete objects if present. Output strictly as a comma-separated list of short phrases (max 40 items). Do not include abstract words.
\end{skeletonbox}
\end{widepromptcard}

\begin{widepromptcard}{pc:mramg-text-json}{2a. Offline: Text-to-Concept Extraction (JSON-Strict)}
\begin{tabularx}{\linewidth}{@{}p{0.13\linewidth}X@{}}
\pcfield{Role}{Extract key entities, methods, datasets, metrics, and core topics from a text chunk for offline indexing.}
\pcfield{Inputs}{Text chunk.}
\pcfield{Output}{Valid JSON array of strings only.}
\pcfield{Gate}{If cannot comply, output \texttt{[]} only.}
\end{tabularx}

\pcsep
\textbf{Hard Constraints.}
\begin{itemize}[leftmargin=1.6em, topsep=1pt, itemsep=1pt, parsep=0pt]
  \item \textbf{Strict JSON}: output ONLY a valid JSON array of strings; no explanation, no extra text.
  \item \textbf{Failure mode}: if cannot comply, output \texttt{[]} only.
\end{itemize}

\pcsep
\textbf{Skeleton.}
\begin{skeletonbox}
\ttfamily\footnotesize\raggedright\sloppy
You are an information extraction engine.\par
Extract key entities, methods, datasets, metrics, and core topics.\par
OUTPUT RULES (must follow):\par
1) Output ONLY a valid JSON array of strings.\par
2) No explanation, no extra text.\par
3) If you cannot comply, output [] only.\par
Example: ["mscoco","flickr30k","transformer","recall@1","map"]\par
\medskip
TEXT:\par
\{text\}
\end{skeletonbox}
\end{widepromptcard}

\begin{widepromptcard}{pc:mramg-text-lines}{2b. Offline: Text-to-Concept Extraction (One-Concept-Per-Line)}
\begin{tabularx}{\linewidth}{@{}p{0.13\linewidth}X@{}}
\pcfield{Role}{Extract key entities, methods, datasets, metrics, and core topics from a text chunk for offline indexing (line-based output).}
\pcfield{Inputs}{Text chunk.}
\pcfield{Output}{Concept list, one concept per line, or \texttt{EMPTY}.}
\pcfield{Gate}{If none, output \texttt{EMPTY}.}
\end{tabularx}

\pcsep
\textbf{Hard Constraints.}
\begin{itemize}[leftmargin=1.6em, topsep=1pt, itemsep=1pt, parsep=0pt]
  \item \textbf{One per line}: one concept per line; no numbering/bullets; no explanation.
  \item \textbf{Empty case}: if none, output \texttt{EMPTY}.
\end{itemize}

\pcsep
\textbf{Skeleton.}
\begin{skeletonbox}
\ttfamily\footnotesize\raggedright\sloppy
Extract key entities, methods, datasets, metrics, and core topics.\par
OUTPUT RULES:\par
1) Output ONLY concepts.\par
2) One concept per line.\par
3) No numbering, no bullets, no explanation.\par
4) If none, output EMPTY.\par
\medskip
TEXT:\par
\{text\}
\end{skeletonbox}
\end{widepromptcard}

\begin{widepromptcard}{pc:mramg-text-tags}{2c. Offline: Text-to-Concept Extraction (Tag-Wrapped)}
\begin{tabularx}{\linewidth}{@{}p{0.13\linewidth}X@{}}
\pcfield{Role}{Extract key entities, methods, datasets, metrics, and core topics from a text chunk for offline indexing (robust tag parsing).}
\pcfield{Inputs}{Text chunk.}
\pcfield{Output}{Concept list wrapped in \texttt{<CONCEPTS>...</CONCEPTS>}.}
\pcfield{Gate}{If none, return \texttt{<CONCEPTS></CONCEPTS>}.}
\end{tabularx}

\pcsep
\textbf{Hard Constraints.}
\begin{itemize}[leftmargin=1.6em, topsep=1pt, itemsep=1pt, parsep=0pt]
  \item \textbf{Tag-only}: return concepts ONLY inside \texttt{<CONCEPTS>...</CONCEPTS>}, one per line.
  \item \textbf{Empty case}: if none, return exactly \texttt{<CONCEPTS></CONCEPTS>}.
\end{itemize}

\pcsep
\textbf{Skeleton.}
\begin{skeletonbox}
\ttfamily\footnotesize\raggedright\sloppy
Extract key entities, methods, datasets, metrics, and core topics.\par
Return concepts inside <CONCEPTS>...</CONCEPTS>, one per line.\par
If none, return <CONCEPTS></CONCEPTS>.\par
\medskip
TEXT:\par
\{text\}
\end{skeletonbox}
\end{widepromptcard}

\begin{widepromptcard}{pc:mramg-query-parsing}{3. Online: Query Parsing for Mixed-Logical Retrieval}
\begin{tabularx}{\linewidth}{@{}p{0.13\linewidth}X@{}}
\pcfield{Role}{Parse user query into positive/negative visual concepts for Neo4j + Milvus retrieval.}
\pcfield{Inputs}{User query text.}
\pcfield{Output}{Strict JSON object with keys \texttt{'positive'} and \texttt{'negative'}.}
\pcfield{Gate}{N/A.}
\end{tabularx}

\pcsep
\textbf{Hard Constraints.}
\begin{itemize}[leftmargin=1.6em, topsep=1pt, itemsep=1pt, parsep=0pt]
  \item \textbf{Visual-verifiable only}: concepts must be concrete and observable in figures.
  \item \textbf{No abstraction}: do NOT output abstract words, e.g., ``efficient'' or ``improve''.
  \item \textbf{Strict JSON}: return valid JSON with keys \texttt{'positive'} and \texttt{'negative'} only.
\end{itemize}

\pcsep
\textbf{Skeleton.}
\begin{skeletonbox}
\ttfamily\footnotesize\raggedright\sloppy
Extract ONLY visual-verifiable concepts for retrieving a scientific figure (arXiv-style).\par
Return strictly valid JSON with keys 'positive' and 'negative'.\par
Positive concepts should be concrete objects/attributes/plot elements/layout cues.\par
Negative concepts are things the figure should NOT contain if explicitly stated.\par
Example: \{"positive": ["line plot","axis","legend"], "negative": ["table"]\}.\par
Query: \{q\_for\_model\}
\end{skeletonbox}
\end{widepromptcard}

\begin{widepromptcard}{pc:mramg-coordinate-answer}{4. Online: Coordinate-Aware Multimodal Answer Generation}
\begin{tabularx}{\linewidth}{@{}p{0.13\linewidth}X@{}}
\pcfield{Role}{Answer using a global view + local slices with normalized coordinates; optionally include retrieved text evidence.}
\pcfield{Inputs}{Global image; local slices with coords \texttt{[x1,y1,x2,y2]} (0--1000); optional texts; question.}
\pcfield{Output}{Free-form answer text.}
\pcfield{Gate}{N/A.}
\end{tabularx}

\pcsep
\textbf{Hard Constraints.}
\begin{itemize}[leftmargin=1.6em, topsep=1pt, itemsep=1pt, parsep=0pt]
  \item \textbf{Coordinate grounding}: use normalized coordinates to reason about spatial layout across slices.
  \item \textbf{Evidence discipline}: do not invent unsupported details.
\end{itemize}

\pcsep
\textbf{Skeleton.}
\begin{skeletonbox}
\ttfamily\footnotesize\raggedright\sloppy
System:\par
You are an expert in visual reasoning. I will provide a global view followed by detailed local slices with their normalized coordinates [x1, y1, x2, y2] (scale 0-1000). Use the coordinates to understand spatial layout.\par
\medskip
User (content order):\par
[Retrieved Text Evidence] (optional):\par
(1) \detokenize{<text>} ...\par
\medskip
[Global View]: \detokenize{<image>}\par
[Local Slices]:\par
Slice [x1, y1, x2, y2]: \detokenize{<image>}\par
...\par
\medskip
Question: \{query\}
\end{skeletonbox}
\end{widepromptcard}

\begin{widepromptcard}{pc:mramg-evidence-rag}{5. Online: Evidence-Grounded RAG Answer Generation}
\begin{tabularx}{\linewidth}{@{}p{0.13\linewidth}X@{}}
\pcfield{Role}{Answer the question using retrieved evidence (texts + images).}
\pcfield{Inputs}{Retrieved text snippets; retrieved images; question.}
\pcfield{Output}{Free-form answer, conservative if evidence is insufficient.}
\pcfield{Gate}{N/A.}
\end{tabularx}

\pcsep
\textbf{Hard Constraints.}
\begin{itemize}[leftmargin=1.6em, topsep=1pt, itemsep=1pt, parsep=0pt]
  \item \textbf{Evidence-only}: use provided evidence to answer; if insufficient, answer conservatively.
  \item \textbf{No hallucination}: avoid adding details not supported by retrieved evidence.
\end{itemize}

\pcsep
\textbf{Skeleton.}
\begin{skeletonbox}
\ttfamily\footnotesize\raggedright\sloppy
System:\par
You are an expert assistant. Use the provided evidence (text and images) to answer the question. If the evidence is insufficient, answer conservatively.\par
\medskip
User (content order):\par
[Retrieved Text Evidence] (optional):\par
(1) \detokenize{<text>} ...\par
\medskip
[Retrieved Images] (optional):\par
Image 1: \detokenize{<image>}\par
...\par
\medskip
Question: \{query\}
\end{skeletonbox}
\end{widepromptcard}

\begin{widepromptcard}{pc:mramg-reranking}{6. Online: Candidate Answer Selection (Reranking)}
\begin{tabularx}{\linewidth}{@{}p{0.13\linewidth}X@{}}
\pcfield{Role}{Select the single best answer among candidates.}
\pcfield{Inputs}{Question; candidate answers indexed from 0 to \(N-1\).}
\pcfield{Output}{Single integer index (0-based) only.}
\pcfield{Gate}{If parsing fails, default index 0.}
\end{tabularx}

\pcsep
\textbf{Hard Constraints.}
\begin{itemize}[leftmargin=1.6em, topsep=1pt, itemsep=1pt, parsep=0pt]
  \item \textbf{Index-only}: return ONLY the index number (0-based) as a single integer.
  \item \textbf{No extra text}: do not output explanations or any additional tokens.
\end{itemize}

\pcsep
\textbf{Skeleton.}
\begin{skeletonbox}
\ttfamily\footnotesize\raggedright\sloppy
You are a strict evaluator. Select the single best answer that directly and correctly answers the question.\par
Return ONLY the index number (0-based) as a single integer.\par
\medskip
Question: \{question\}\par
\medskip
Candidates:\par
\{items\}
\end{skeletonbox}
\end{widepromptcard}

\subsubsection{MRAMG Qualitative Examples}

\begin{widepromptcard}{pc:mramg-example-laser}{Example E1: LaserGuider (physical backdoor attack stages)}
\begin{tabularx}{\linewidth}{@{}p{0.13\linewidth}X@{}}
\pcfield{Role}{Qualitative visualization of an online inference instance (question, retrieval, grounding, and output).}
\pcfield{Inputs}{Question; retrieved doc/image IDs; selected patch coordinates; model output; reference answer; evaluation signals.}
\pcfield{Output}{A compact, prompt-style trace for appendix presentation.}
\pcfield{Gate}{N/A.}
\end{tabularx}

\pcsep
\textbf{Hard Constraints.}
\begin{itemize}[leftmargin=1.6em, topsep=1pt, itemsep=1pt, parsep=0pt]
  \item \textbf{No chain-of-thought}: we report only a high-level trace (actions and used evidence), without hidden reasoning steps.
  \item \textbf{Verbatim IDs}: document and image identifiers are shown exactly as logged.
  \item \textbf{Layout safety}: bracketed markers are wrapped to avoid \LaTeX{} optional-argument parsing.
\end{itemize}

\pcsep
\textbf{Skeleton.}
\begin{skeletonbox}
\ttfamily\scriptsize\raggedright\sloppy
\textbf{Question:} In LaserGuider how does the adversary embed and trigger a backdoor in DNNs using laser-based techniques, and what are the key stages of the attack?\par
\textbf{Retrieved Docs (top-5):} arxiv\_doc\_60048, arxiv\_doc\_60099, arxiv\_doc\_60090, arxiv\_doc\_60096, arxiv\_doc\_60047\par
\textbf{GT Docs (top-5):} arxiv\_doc\_60047\par
\textbf{Top Images (top-4):} 2412\_03993v1\_0, 2106\_09282v1\_0, 2312\_04512v2\_0, 2312\_04512v2\_1\par
\textbf{Used Patches:} vqa [0,0,875,1000]; vqa [125,0,1000,1000]\par
\textbf{Model Answer:} In the context of LaserGuider, the adversary employs a two-stage process to embed and trigger a backdoor in DNNs using laser-based techniques: Backdoor Embedding: (i) select images and define a digital trigger design, (ii) create a poisoned dataset, and (iii) train the DNN to associate the trigger with a target behavior. Backdoor Triggering: (i) project the corresponding physical trigger via a laser during inference, (ii) the model recognizes the trigger pattern, and (iii) the backdoor behavior is activated.\par
\textbf{Reference Answer (GT):} The attack proceeds in two stages: (1) backdoor embedding by poisoning training data using a designed digital laser-trigger $\delta$ and retraining so the model learns a trigger-to-target association; (2) backdoor triggering at inference by projecting a physically realizable laser pattern that corresponds to $\delta$, causing the model to activate the malicious target behavior.\par
\textbf{Trace (high-level):} Retrieve top-K text+image evidence; select local visual patches using normalized coordinates; generate an answer constrained to retrieved evidence; evaluate overlap with the reference answer.\par
\textbf{Outcome:} token\_F1=0.432, ROUGE-L=0.225, BLEU-4=0.059; doc\_hit@5=1, img\_hit@10=1\par
\end{skeletonbox}
\end{widepromptcard}

\begin{widepromptcard}{pc:mramg-example-civrealm}{Example E2: CivRealm (key features)}
\begin{tabularx}{\linewidth}{@{}p{0.13\linewidth}X@{}}
\pcfield{Role}{Qualitative visualization of an online inference instance (question, retrieval, grounding, and output).}
\pcfield{Inputs}{Question; retrieved doc/image IDs; selected patch coordinates; model output; reference answer; evaluation signals.}
\pcfield{Output}{A compact, prompt-style trace for appendix presentation.}
\pcfield{Gate}{N/A.}
\end{tabularx}

\pcsep
\textbf{Hard Constraints.}
\begin{itemize}[leftmargin=1.6em, topsep=1pt, itemsep=1pt, parsep=0pt]
  \item \textbf{No chain-of-thought}: we report only a high-level trace (actions and used evidence), without hidden reasoning steps.
  \item \textbf{Verbatim IDs}: document and image identifiers are shown exactly as logged.
  \item \textbf{Layout safety}: bracketed markers are wrapped to avoid \LaTeX{} optional-argument parsing.
\end{itemize}

\pcsep
\textbf{Skeleton.}
\begin{skeletonbox}
\ttfamily\scriptsize\raggedright\sloppy
\textbf{Question:} What are the features of CivRealm?\par
\textbf{Retrieved Docs (top-5):} arxiv\_doc\_60048, arxiv\_doc\_60011, arxiv\_doc\_60017, arxiv\_doc\_60096, arxiv\_doc\_60091\par
\textbf{GT Docs (top-5):} arxiv\_doc\_60096\par
\textbf{Top Images (top-4):} 2403\_16073v3\_3, 2403\_16073v3\_4, 2401\_10568v2\_0, 2403\_16073v3\_2\par
\textbf{Used Patches:} vqa [0,0,1000,875]; vqa [0,125,1000,1000]\par
\textbf{Model Answer:} CivRealm is an open-ended testbed for decision-making agents, featuring imperfect information, stochastic dynamics, multiple victory paths, a dynamic game space, multi-agent interactions (including alliances/self-play), general-sum outcomes, and communication/diplomacy mechanisms.\par
\textbf{Reference Answer (GT):} CivRealm is an open-ended decision-making environment with imperfect information, stochastic events, multiple victory conditions, a dynamically changing state/action space, multi-agent interactions (including alliances), and communication/diplomacy that can leverage natural language.\par
\textbf{Trace (high-level):} Retrieve top-K text+image evidence; select local visual patches using normalized coordinates; generate an answer constrained to retrieved evidence; evaluate overlap with the reference answer.\par
\textbf{Outcome:} token\_F1=0.253, ROUGE-L=0.182, BLEU-4=0.040; doc\_hit@5=1, img\_hit@10=1\par
\end{skeletonbox}
\end{widepromptcard}

\subsubsection{MRAG Prompt Templates}

\begin{widepromptcard}{pc:mrag-full-image-caption}{1. Offline: Full-Image Concept Caption (for Retrieval Indexing)}
\begin{tabularx}{\linewidth}{@{}>{\bfseries}p{0.13\linewidth}X@{}}
Role & Extract key visual concepts from the full image for offline retrieval indexing.\\
Inputs & One image, global view, loaded from \texttt{img\_path}.\\
Output & Comma-separated list of key visual concepts, using short phrases.\\
Gate & N/A.\\
\end{tabularx}

\pcsep
\textbf{Hard Constraints.}
\begin{itemize}[leftmargin=1.6em, topsep=1pt, itemsep=1pt, parsep=0pt]
  \item \textbf{Multi-facet coverage}: include \emph{category/identity}, \emph{shape/parts/structure}, \emph{visual attributes}, and \emph{state/condition} only if visible.
  \item \textbf{Format}: output strictly as a comma-separated list; no numbering/bullets; no extra text.
  \item \textbf{Concrete phrases}: keep items short and visual-verifiable; avoid generic words such as \emph{characteristics} or \emph{feature}.
\end{itemize}

\pcsep
\textbf{Skeleton.}
\begin{skeletonbox}
\ttfamily\footnotesize\raggedright\sloppy
[image]\par
List key visual concepts for retrieval as a comma-separated list. Cover multiple facets: category/identity, shape/parts/structure, visual attributes such as color, texture, spots, or mold, and state/condition if visible. Avoid generic words like 'characteristics' or 'feature'.
\end{skeletonbox}
\end{widepromptcard}

\begin{widepromptcard}{pc:mrag-concept-vqa}{2. Offline: Concept VQA (Binary Presence Check for Gating)}
\begin{tabularx}{\linewidth}{@{}p{0.13\linewidth}X@{}}
\pcfield{Role}{Query the model with a binary existence question for each candidate concept to obtain concept-level signals for gating.}
\pcfield{Inputs}{One image, global view, plus one candidate concept string \texttt{c}.}
\pcfield{Output}{Exactly \texttt{Yes} or \texttt{No}.}
\pcfield{Gate}{Otsu gate over concept-level scores to split candidate concepts into positive vs.\ negative sets.}
\end{tabularx}

\pcsep
\textbf{Hard Constraints.}
\begin{itemize}[leftmargin=1.6em, topsep=1pt, itemsep=1pt, parsep=0pt]
  \item \textbf{Binary-only}: output must be exactly \texttt{Yes} or \texttt{No}; no punctuation, explanation, or extra tokens.
  \item \textbf{Fixed phrasing}: do not paraphrase the question; only substitute \texttt{c}.
  \item \textbf{Visual-verifiable}: answer based only on what is visible in the image.
\end{itemize}

\pcsep
\textbf{Skeleton.}
\begin{skeletonbox}
\ttfamily\footnotesize\raggedright\sloppy
[image]\par
Is there a \{c\} in this image? Answer Yes or No.
\end{skeletonbox}
\end{widepromptcard}

\begin{widepromptcard}{pc:mrag-query-parsing}{3. Online: Query Parsing into Retrieval Concepts (\texttt{positive}/\texttt{negative} JSON)}
\begin{tabularx}{\linewidth}{@{}>{\bfseries}p{0.13\linewidth}X@{}}
Role & Parse the user question and the main image if provided into retrieval-oriented visual concepts for Milvus/Neo4j retrieval.\\
Inputs & User query text \texttt{query}; optional main image.\\
Output & Strict JSON only: \texttt{\{"positive":[...], "negative":[...]\}}.\\
Gate & N/A.\\
\end{tabularx}

\pcsep
\textbf{Hard Constraints.}
\begin{itemize}[leftmargin=1.6em, topsep=1pt, itemsep=1pt, parsep=0pt]
  \item \textbf{Use both modalities if available}: use both the question and the main image if an image is provided; otherwise use the question only.
  \item \textbf{Concrete and visual}: concepts must be visual-verifiable; avoid abstract placeholders.
  \item \textbf{Short phrases}: each concept should be 1--3 words and retrieval-friendly.
  \item \textbf{Negative rule}: include negatives only if explicitly stated as not wanted or unlikely; otherwise output \texttt{[]}.
  \item \textbf{Strict JSON}: output valid JSON only; no prose, markdown, or extra keys.
\end{itemize}

\pcsep
\textbf{Skeleton.}
\begin{skeletonbox}
\ttfamily\scriptsize\raggedright\sloppy
[optional image]\\
Extract key visual concepts for image retrieval using BOTH the question and the main image (if provided).\\
\medskip
Consider multiple facets (balanced):\\
1) Category/identity (e.g., fruit type, object class).\\
2) Shape/parts/structure (e.g., sliced, cross-section, seeds, stem).\\
3) Visual attributes (color, texture, surface patterns, spots, mold/fuzz, moisture).\\
4) State/condition (fresh/ripe/oxidized/rotting) only if mentioned or visible.\\
\medskip
Guidelines:\\
- Use concrete, observable phrases.\\
- Avoid generic words like 'characteristics', 'feature', 'aspect'.\\
- Keep each concept short (1-3 words) and specific.\\
\medskip
Rules:\\
- 'positive': concepts that SHOULD appear in the target image.\\
- 'negative': concepts explicitly stated as NOT wanted OR explicitly stated as unlikely in the question.\\
  If no negative concepts, use [].\\
\medskip
Output ONLY valid JSON and nothing else.\\
Format: \{"positive": [...], "negative": [...]\}\\
\medskip
Question: \{query\}\\
\medskip
JSON:
\end{skeletonbox}
\end{widepromptcard}

\begin{widepromptcard}{pc:mrag-baseline-mcq}{4. Online: Baseline MCQ Inference (Main Image Only; No Retrieval)}
\begin{tabularx}{\linewidth}{@{}>{\bfseries}p{0.13\linewidth}X@{}}
Role & Answer a multiple-choice question using only the MAIN IMAGE, with strictly formatted output.\\
Inputs & Main image + question \texttt{\{question\}} + options \texttt{\{A,B,C,D\}}.\\
Output & Exactly one line: \texttt{Answer: X}, where \texttt{X} $\in$ \{A,B,C,D\}.\\
Gate & N/A.\\
\end{tabularx}

\pcsep
\textbf{Hard Constraints.}
\begin{itemize}[leftmargin=1.6em, topsep=1pt, itemsep=1pt, parsep=0pt]
  \item \textbf{Main-image only}: do not assume external knowledge or retrieved context.
  \item \textbf{No explanation}: do not output reasoning, evidence, or uncertainty.
  \item \textbf{Strict format}: output must be exactly \texttt{Answer: X} with a single option letter.
  \item \textbf{Forced choice}: choose the single most likely option based on the main image.
\end{itemize}

\pcsep
\textbf{Skeleton.}
\begin{skeletonbox}
\ttfamily\scriptsize\raggedright\sloppy
[main image]\\
You are answering a multiple choice question about the MAIN IMAGE.\\
\medskip
\#\# Question: \{question\}\\
\medskip
\#\# Options:\\
A) \{A\}\\
B) \{B\}\\
C) \{C\}\\
D) \{D\}\\
\medskip
\#\# Instructions:\\
1. Analyze the MAIN IMAGE carefully to answer the question.\\
2. You MUST choose the MOST LIKELY correct option based on the main image.\\
3. Do NOT express uncertainty, doubt, or provide explanations.\\
4. Output ONLY the answer in the exact format below.\\
\medskip
\#\# Output Format (STRICT):\\
Answer: X\\
where X is one of A, B, C, or D.\\
\medskip
CRITICAL: Do NOT include any text, explanation, reasoning, or uncertainty after 'Answer: X'. Just the answer.
\end{skeletonbox}
\end{widepromptcard}

\begin{widepromptcard}{pc:mrag-rag-mcq}{5. Online: RAG MCQ Inference (Retrieved Images + Relationship Context)}
\begin{tabularx}{\linewidth}{@{}>{\bfseries}p{0.13\linewidth}X@{}}
Role & Answer a multiple-choice question using the MAIN IMAGE plus RETRIEVED IMAGES, additionally conditioned on explicit image-relationship context.\\
Inputs & Main image + retrieved images + relationship context \texttt{\{relationship\_context\}} + question \texttt{\{question\}} + options \texttt{\{A,B,C,D\}}.\\
Output & Exactly one line: \texttt{Answer: X}, where \texttt{X} $\in$ \{A,B,C,D\}.\\
Gate & N/A; relationships already computed.\\
\end{tabularx}

\pcsep
\textbf{Hard Constraints.}
\begin{itemize}[leftmargin=1.6em, topsep=1pt, itemsep=1pt, parsep=0pt]
  \item \textbf{Main image first}: examine the main image before consulting retrieved images and relationships.
  \item \textbf{Relationship-aware}: treat \texttt{\{relationship\_context\}} as auxiliary structure; do not hallucinate relationships not provided.
  \item \textbf{No explanation}: output no reasoning, evidence, or uncertainty.
  \item \textbf{Strict format}: output must be exactly \texttt{Answer: X} with a single option letter.
\end{itemize}

\pcsep
\textbf{Skeleton.}
\begin{skeletonbox}
\ttfamily\scriptsize\raggedright\sloppy
[main image][retrieved images]\\
You are answering a multiple choice question using RETRIEVAL-AUGMENTED reasoning.\\
\medskip
\#\# Image Structure:\\
- MAIN IMAGE: The question image that you need to analyze.\\
- RETRIEVED IMAGES: Similar/related images retrieved from a knowledge base. These images contain relevant visual patterns, concepts, or examples that can help you better understand the main image and answer the question.\\
\medskip
\#\# Image Relationships:\\
\{relationship\_context\}\\
\medskip
\#\# Question: \{question\}\\
\medskip
\#\# Options:\\
A) \{A\}\\
B) \{B\}\\
C) \{C\}\\
D) \{D\}\\
\medskip
\#\# Instructions (RETRIEVAL-AUGMENTED):\\
1. First, carefully examine the MAIN IMAGE to understand what it shows.\\
2. Then, analyze the RETRIEVED IMAGES to identify relevant visual patterns, concepts, or similar cases that relate to the question.\\
3. Use the retrieved images to enhance your understanding of the main image - they may show similar objects, states, transformations, or provide context that helps interpret the main image.\\
4. Consider the relationships between images (similarity links) to understand how they connect to the main image.\\
5. Synthesize information from both the main image and retrieved images to determine the MOST LIKELY correct answer.\\
6. Do NOT express uncertainty, doubt, or provide explanations.\\
7. Output ONLY the answer in the exact format below.\\
\medskip
\#\# Output Format (STRICT):\\
Answer: X\\
where X is one of A, B, C, or D.\\
\medskip
CRITICAL: Do NOT include any text, explanation, reasoning, or uncertainty after 'Answer: X'. Just the answer.
\end{skeletonbox}
\end{widepromptcard}

\subsubsection{MRAG Qualitative Visualizations}

\begin{widepromptcard}{pc:norag-q6}{NoRAG (Baseline) Inference Visualization: Sample \#6 (Biological, Transformative)}
\begin{tabularx}{\linewidth}{@{}>{\bfseries}p{0.13\linewidth}X@{}}
Role & Answer a multiple-choice question using only the MAIN IMAGE (no retrieval), with strictly formatted output.\\
Inputs & (1) MAIN IMAGE (2) Question + four options (A--D).\\
Output & Exactly one line: \texttt{Answer: X}, where \texttt{X} $\in$ \{A,B,C,D\}.\\
Gate & N/A.\\
\end{tabularx}

\pcsep
\textbf{Visualized Inputs (what the model sees).}
\begin{center}
\includegraphics[width=0.1\linewidth,height=0.1\textheight]{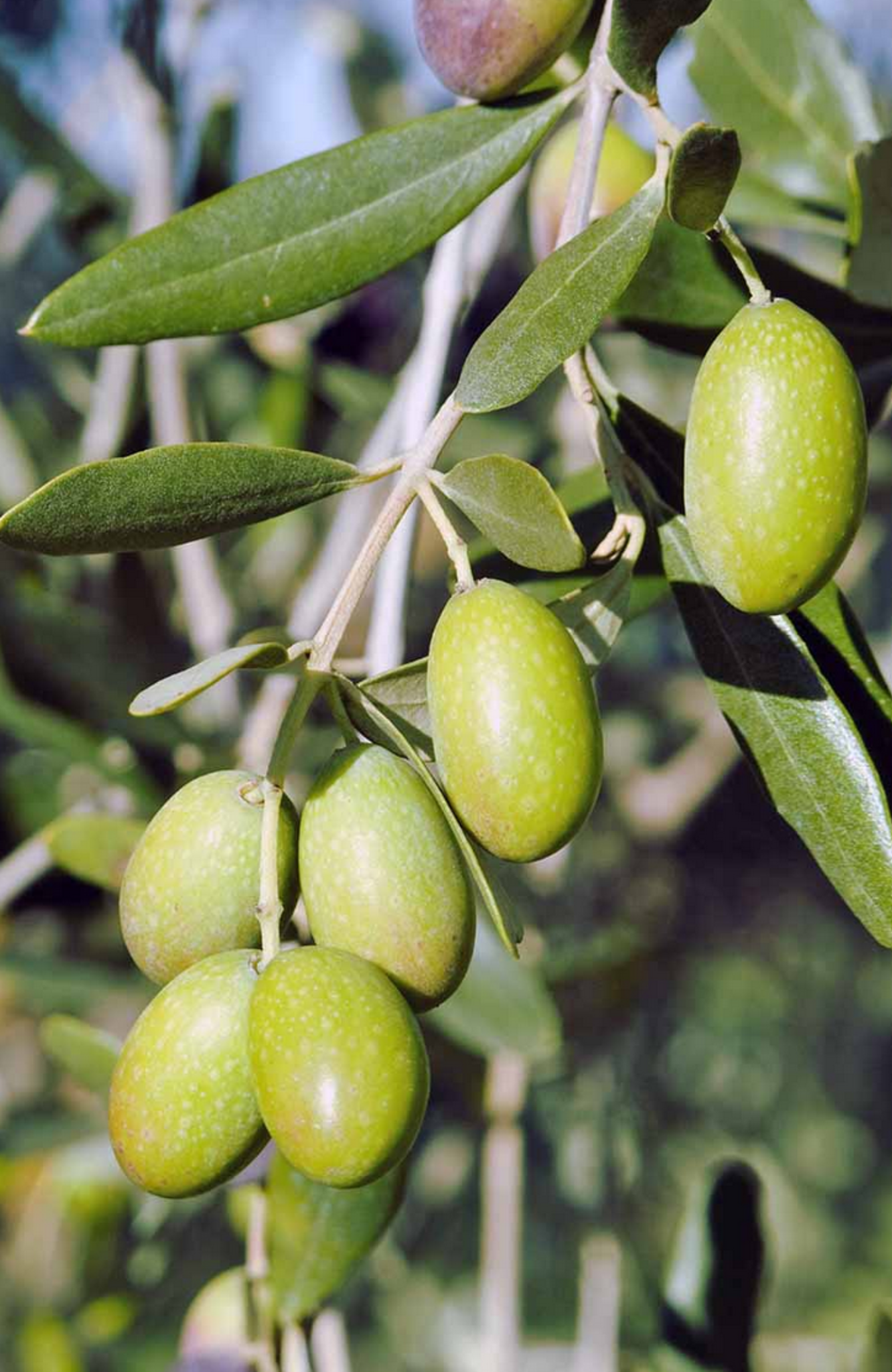}
\end{center}
\begin{skeletonbox}
\ttfamily\small
You are answering a multiple choice question about the MAIN IMAGE.\\
\medskip
\#\# Question: Among these features, which one is unlikely for this fruit once it undergoes oxidation?\\
\medskip
\#\# Options:\\
A) Its color changes to a light brown.\\
B) Its skin remains smooth and shiny.\\
C) A blueish-green mold forms on its surface.\\
D) White fuzzy mold grows on it.\\
\medskip
\#\# Instructions:\\
1. Analyze the MAIN IMAGE carefully to answer the question.\\
2. You MUST choose the MOST LIKELY correct option based on the main image.\\
3. Do NOT express uncertainty, doubt, or provide explanations.\\
4. Output ONLY the answer in the exact format below.\\
\medskip
\#\# Output Format (STRICT):\\
Answer: X\\
where X is one of A, B, C, or D.\\
\medskip
CRITICAL: Do NOT include any text, explanation, reasoning, or uncertainty after 'Answer: X'. Just the answer.
\end{skeletonbox}

\pcsep
\textbf{Observed Output (this run).}
\begin{skeletonbox}
\ttfamily\small
Answer: B
\end{skeletonbox}
\end{widepromptcard}

\begin{widepromptcard}{pc:rag-q6}{RAG (Full Retrieval) Inference Visualization: Sample \#6 (Biological, Transformative)}
\begin{tabularx}{\linewidth}{@{}>{\bfseries}p{0.13\linewidth}X@{}}
Role & Answer a multiple-choice question using the MAIN IMAGE plus RETRIEVED IMAGES as auxiliary visual evidence, with strictly formatted output.\\
Inputs & (1) MAIN IMAGE: \texttt{q6\_main.png}; (2) RETRIEVED IMAGES: \texttt{q6\_r1.png--q6\_r5.png}; (3) Question + four options (A--D).\\
Output & Exactly one line: \texttt{Answer: X}, where \texttt{X} $\in$ \{A,B,C,D\}.\\
Gate & Retrieval is performed beforehand (e.g., Milvus top-$k$); this card visualizes the final model input.\\
\end{tabularx}

\pcsep
\textbf{Visualized Inputs (what the model sees).}

\begin{center}
\small
\textbf{Left: MAIN IMAGE \quad\quad Right: RETRIEVED IMAGES (top-$k$)}
\end{center}

\begin{center}
\begin{tabular}{cccccc}
\includegraphics[width=0.1\linewidth,height=0.1\textheight]{fig/q6_main.png} &
\includegraphics[width=0.1\linewidth,height=0.1\textheight]{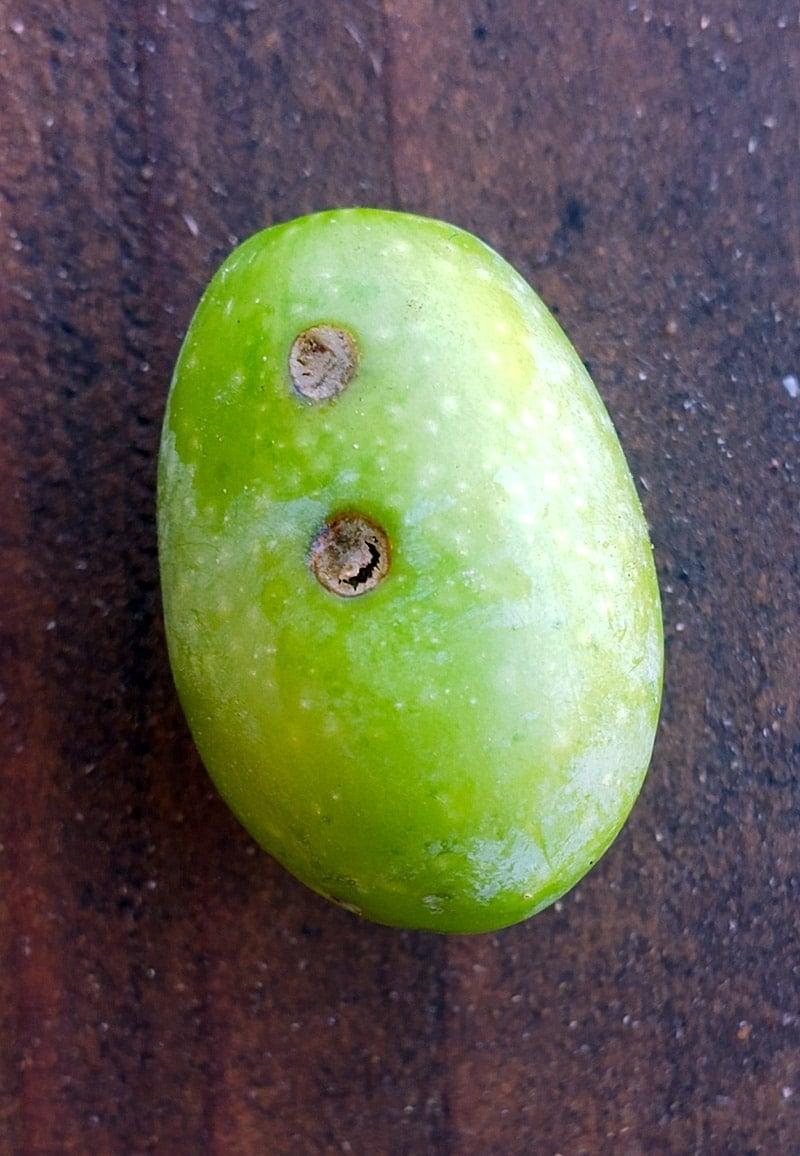} &
\includegraphics[width=0.1\linewidth,height=0.1\textheight]{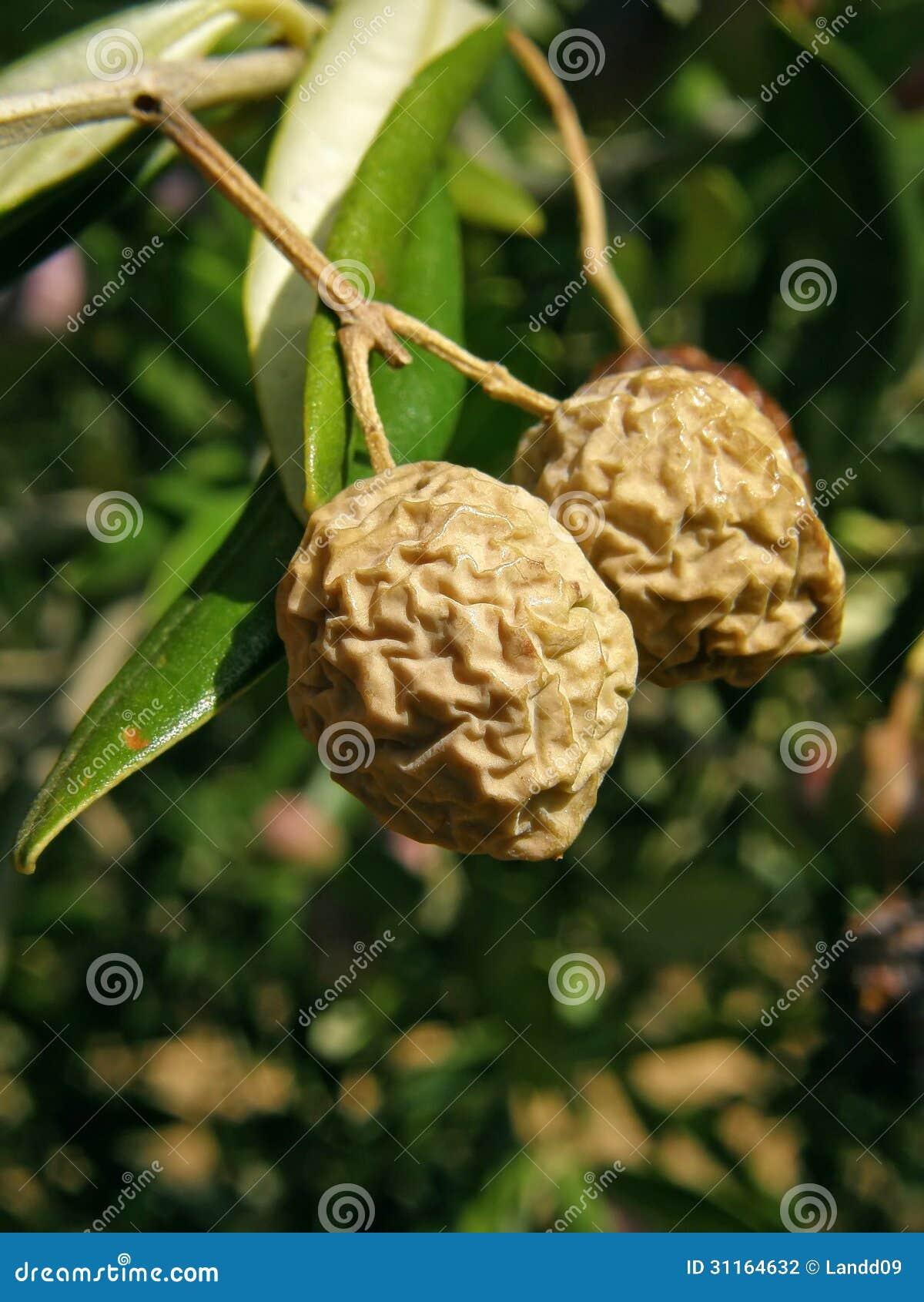} &
\includegraphics[width=0.1\linewidth,height=0.1\textheight]{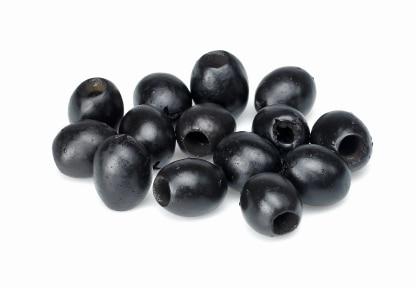} &
\includegraphics[width=0.1\linewidth,height=0.1\textheight]{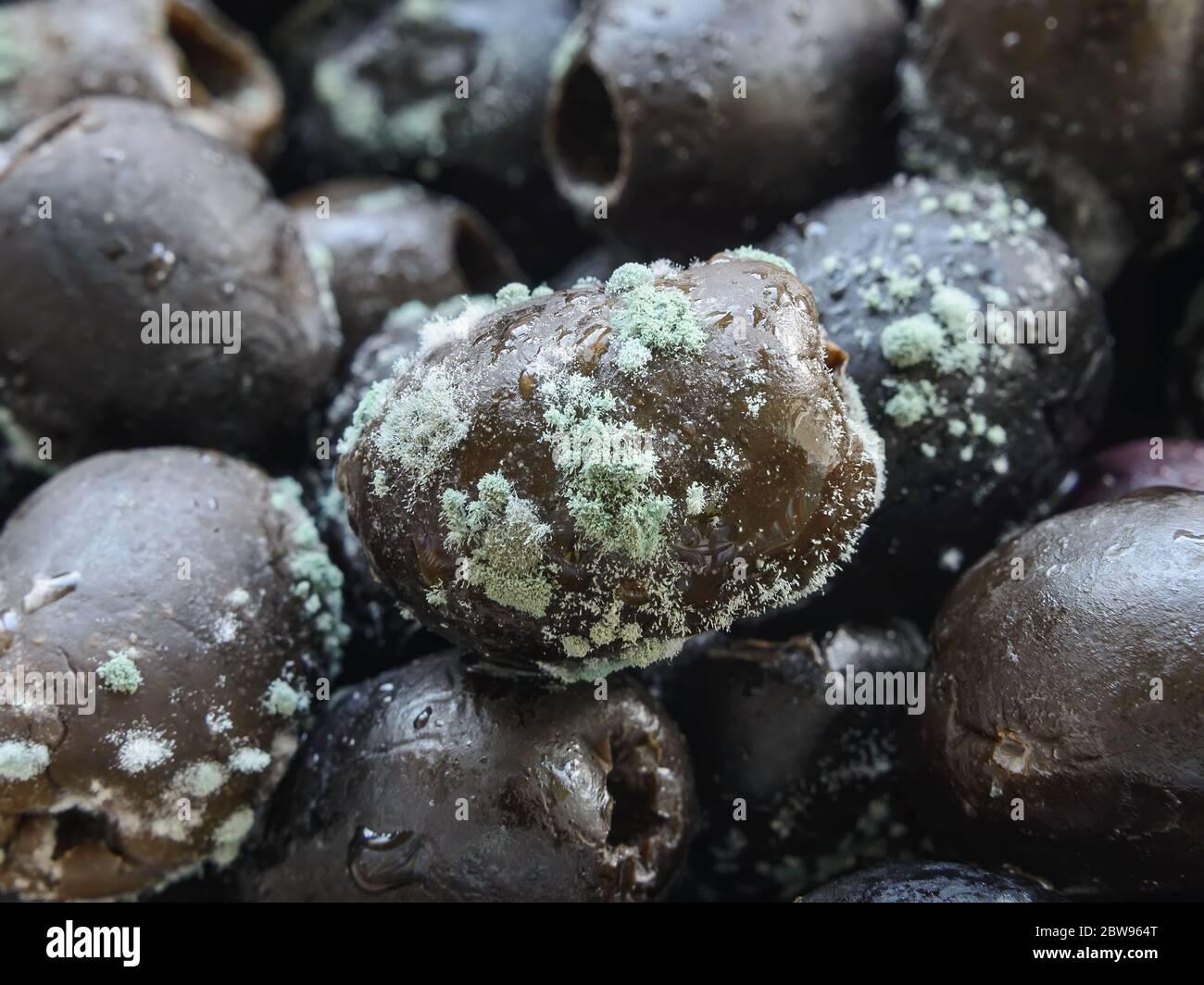} &
\includegraphics[width=0.1\linewidth,height=0.1\textheight]{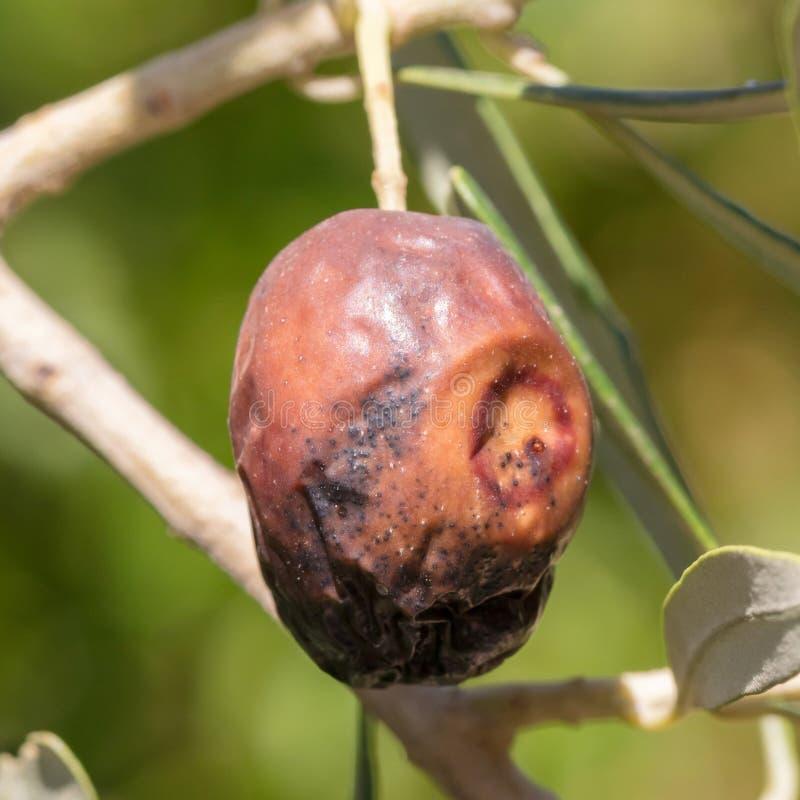} \\
\end{tabular}
\end{center}

\begin{skeletonbox}
\ttfamily\small
You are answering a multiple choice question using RETRIEVAL-AUGMENTED reasoning.\\
\medskip
\#\# Image Structure:\\
- MAIN IMAGE: The question image that you need to analyze.\\
- RETRIEVED IMAGES: Similar/related images retrieved from a knowledge base. These images contain relevant visual patterns, concepts, or examples that can help you better understand the main image and answer the question.\\
\medskip
\#\# Question: Among these features, which one is unlikely for this fruit once it undergoes oxidation?\\
\medskip
\#\# Options:\\
A) Its color changes to a light brown.\\
B) Its skin remains smooth and shiny.\\
C) A blueish-green mold forms on its surface.\\
D) White fuzzy mold grows on it.\\
\medskip
\#\# Instructions (RETRIEVAL-AUGMENTED):\\
1. First, carefully examine the MAIN IMAGE to understand what it shows.\\
2. Then, analyze the RETRIEVED IMAGES to identify relevant visual patterns, concepts, or similar cases that relate to the question.\\
3. Use the retrieved images to enhance your understanding of the main image - they may show similar objects, states, transformations, or provide context that helps interpret the main image.\\
4. Synthesize information from both the main image and retrieved images to determine the MOST LIKELY correct answer.\\
5. Do NOT express uncertainty, doubt, or provide explanations.\\
6. Output ONLY the answer in the exact format below.\\
\medskip
\#\# Output Format (STRICT):\\
Answer: X\\
where X is one of A, B, C, or D.\\
\medskip
CRITICAL: Do NOT include any text, explanation, reasoning, or uncertainty after 'Answer: X'. Just the answer.
\end{skeletonbox}

\pcsep
\textbf{Observed Output (this run).}
\begin{skeletonbox}
\ttfamily\small
Answer: B
\end{skeletonbox}
\end{widepromptcard}

\end{document}